\documentclass[twocolumn]{article}

\usepackage[utf8]{inputenc}
\usepackage[T1]{fontenc}
\usepackage{lmodern}
\usepackage{authblk}
\usepackage{hyperref}
\usepackage{url}
\usepackage{booktabs}
\usepackage{amsfonts}
\usepackage{nicefrac}
\usepackage{microtype}
\usepackage{subfiles}
\usepackage{amsmath}
\usepackage{amssymb}
\usepackage{pifont}
\usepackage{graphicx}
\usepackage{bbm}
\usepackage{subcaption}
\usepackage{caption}
\usepackage{wrapfig}
\usepackage{natbib}
\usepackage{mathtools}
\mathtoolsset{showonlyrefs=true}
\usepackage{upgreek}
\usepackage{xcolor}
\usepackage{wrapfig}

\usepackage{amsthm}

\makeatletter
\renewcommand\AB@affilsepx{ \protect\Affilfont}
\makeatother

\newtheorem{theorem}{Theorem}
\newtheorem*{theorem*}{Theorem}
\newtheorem{prop}{Property}
\newtheorem*{prop*}{Property}

\newtheorem{cor}{Corollary}

\hypersetup{colorlinks=true,,allcolors=[rgb]{0,0,0.5}}

\DeclareMathOperator*{\argmax}{arg\,max}

\newcommand{\eqdef}{\mathrel{\mathop:}=}
\newcommand{\norm}[1]{\left\lVert#1\right\rVert}

\newcommand*\samethanks[1][\value{footnote}]{\footnotemark[#1]}

\usepackage{balance} %

\title{Lazy-MDPs: Towards Interpretable Reinforcement Learning \\ by Learning When to Act}
\date{}

\author[1]{Alexis Jacq\thanks{Equal contribution.}}
\author[1, 2, 3]{Johan Ferret\samethanks}
\author[1]{Olivier Pietquin}
\author[1]{Matthieu Geist}
\affil[1]{Google Research, Brain Team}
\affil[2]{Inria, Scool Team \authorcr}
\affil[3]{CRIStAL, CNRS, Université de Lille}

\begin{document}

\onecolumn

\maketitle

\begin{abstract}
    Traditionally, Reinforcement Learning (RL) aims at deciding \textit{how to act} optimally for an artificial agent. We argue that deciding \emph{when to act} is equally important. As humans, we drift from default, instinctive or memorized behaviors to focused, thought-out behaviors when required by the situation. To enhance RL agents with this aptitude, we propose to augment the standard Markov Decision Process and make a new mode of action available: %
    being \textit{lazy}, which defers decision-making to a default policy. 
    In addition, we penalize non-lazy actions in order to encourage minimal effort and have agents focus on critical decisions only. We name the resulting formalism \emph{lazy-MDPs}. We study the theoretical properties of lazy-MDPs, expressing value functions and characterizing optimal solutions. 
    Then we empirically demonstrate that policies learned in lazy-MDPs generally come with a form of interpretability: by construction, they show us the states where the agent takes control over the default policy. We deem those states and corresponding actions important since they explain the difference in performance between the default and the new, lazy policy.   
    With suboptimal policies as default (pretrained or random), we observe that agents are able to get competitive performance in Atari games while only taking control in a limited subset of states.
\end{abstract}

\section{Introduction}

Decision-making is about providing answers to a standard question: \textit{"how to act?"}. While Markov Decision Processes (MDPs)~\citep{puterman1994mdp} provide the canonical formalism to ask this question, Reinforcement Learning (RL) provides algorithms to answer it. In this work, we study a different question: \textit{\textbf{"when and how to act?"}}. There are several motivations for this particular question. First, in many tasks there is only a handful of states that are critical and require complex decision-making, while in other states the action has less impact than the own dynamic of the MDP (for example, the orientation of a falling piece in Tetris has no importance until it reaches the floor). Another motivation is that of learning on top of an existing policy: one might be able to learn a better policy by learning when to take control over this default policy (this is the case in many human-robot interactions~\citep{meresht2020learning}, for example self-driving cars that would let the human drive except in critical situations, robots assisting surgery, etc). Default policies can take arbitrary forms: controllers, handcrafted policies, programs, and many others.    

\begin{wrapfigure}{r}{0.5\textwidth}
  \begin{center}
    \includegraphics[width=\linewidth]{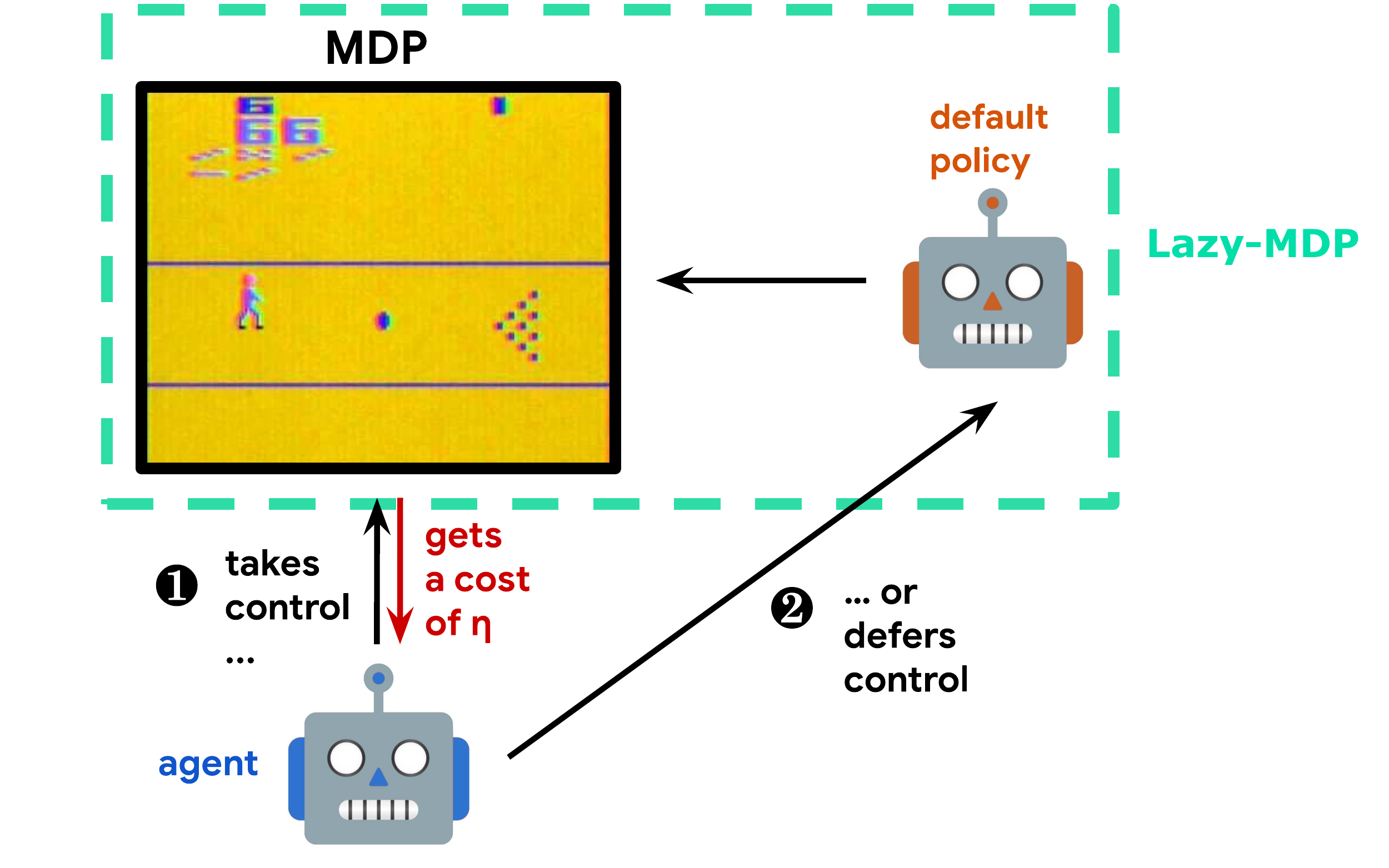}
  \end{center}
  \caption{High-level overview of a lazy-MDP.}
  \label{fig:intuition}
\end{wrapfigure}

To study this alternative question, we need an alternative formalism. Instead of starting from scratch, we propose to augment the existing MDP framework (see Fig.~\ref{fig:intuition}): we extend the action space with a novel action, the \textit{lazy} action; and we modify the reward function to penalize agents when they take control (\textit{i.e.} pick an action from the original action space). Choosing the \textit{lazy} action defers the decision-making process to a default policy. Augmenting the MDP framework means that we can take any decision-making problem that can be expressed as an MDP and turn it into a lazy-MDP. Since defaulting is a discrete action, we chose to focus on discrete actions setting for simplicity, but everything could be adapted to continuous control (for instance using two actors, a default one and a learned one; and one critic for each).

Lazy-MDPs have interesting properties for interpretability: states where the policy diverges from the default hold information.
In more details, we leverage the statewise differences between the default policy and the new, lazy-policy to make sense of what is needed to get performance improvement with respect to the default policy (under arbitrary default policies) or to make sense of the overall task (under specific default policies).
An important point we want to highlight is that the type of interpretability we consider here is different from explanations~\citep{molnar2020interpretable}. Explanations, in the context of RL, would bring answers to "why" questions (either about agent behavior or the importance of actions), which is not what we tackle here. 

Our contributions are the following: 1) we propose a novel formalism called \textit{lazy-MDP} that provides modified decision-making problems where agents have to learn when and how to act, 2) we study lazy-MDPs from a theoretical point of view and prove that we can characterize optimality, which depends on the third-party policy and the value of the penalty for taking control, and 3) we study lazy-MDPs empirically, showing that they lead to learning an interpretable partition of the states. We also study how making control less frequent (by increasing the penalty) impacts the score of agents. In hard exploration tasks, reducing the frequency of controls can even lead to improved performance.

\section{Related work}

\looseness=-1
While the idea of constraining policies to adopt a default behavior as often as possible in MDPs is to the best of our knowledge novel, it lives at the crossroads of several subfields of RL reviewed here. %

\textbf{Residual RL.}
Residual approaches~\citep{silver2018residual, johannink2019residual} consist in learning a residual policy, whose action is added to that of a base policy to get the resulting action. By its nature, residual RL is restricted to continuous control problems, where the sum of two actions is still a valid action. In contrast, the lazy-MDP abstraction is applicable to discrete control problems.

\textbf{Exploration-conscious RL.}
Here we discuss related augmented MDPs. \citet{shani2019exploration} show that exploration-conscious RL~\citep{van2009theoretical} can be solved via a surrogate MDP, where the dynamics are obtained by a linear interpolation of the dynamics induced by the current policy and those induced by a fixed policy; same for rewards. This amounts to learning a policy that is optimal given that the actual behavior is a fixed mixture of that policy and a base one. In comparison, the policies learned in lazy-MDPs are more controllable in the sense that they can switch between the base policy and a learned policy on the basis of states, and serve the subtly different purpose of learning when and how to act.      

\textbf{Interpretable RL.}
There are several types of interpretability studied in the literature. 
Many works try to quantify the influence of parts of the inputs on the decisions of the agents. This can be done via post-hoc gradient methods, using actual policy gradients~\citep{wang2015dueling, zahavy2016graying} or finite-difference estimates~\citep{greydanus2018visualizing, puri2019explain}, coupled with saliency maps as visualizations. 
Another way to do so is by training ad-hoc interpretable models~\citep{liu2018toward, coppens2019distilling} or programs~\citep{verma2018programatically, verma2019imitation} to mimic the non-interpretable models used as policy networks. 
Others try to make sense of the representations learned by agents, using dimensionality reduction techniques~\citep{zahavy2016graying} or state aggregation methods~\citep{topin2019generation}. Another focus is to select trajectories that are representative of the overall behavior of the RL agent~\citep{amir2018highlights, amitai2021disagreement}. 
Finally, some works try to recover the approximate preferences of the agent, under the form of a reward function, or coefficients for a known reward decomposition~\citep{juozapaitis2019explainable, bica2021learning}. 
While interesting on their own, none of the mentioned works explicitly tackle the question of identifying states that are crucial to the decision-making process. The action-gap~\citep{bellemare2016increasing} and importance advising~\citep{torrey2013teaching} are quantities that hint at this aspect. As we show in Sec.~\ref{subsection:interpretability}, both suffer from several shortcomings compared to the proposed method.

\textbf{Credit assignment in RL.}
Temporal credit assignment consists in associating specific actions to specific results (\textit{i.e.} task success or high returns). Existing approaches complement or modify RL algorithms by either decomposing observed returns as the sum of redistributed rewards along observed trajectories~\citep{arjona2019rudder, ferret2019self, hung2019optimizing, raposo2021synthetic} or incorporating hindsight information into the RL process~\citep{harutyunyan2019hindsight, ferret2021self, mesnard2021counterfactual}. Our approach is related but differs in several points: it is tied to (and aims at making sense of) performance improvements instead of outcomes, and it comes from an abstraction over MDPs (which is non-parametric).

\looseness=-1
\textbf{Temporal abstractions in RL.}
Options are common temporal abstractions in RL~\citep{sutton1999between, precup2000temporal, bacon2017option, barreto2019option}. They consist in triples $(\mathcal{I}, \pi, \beta)$ where $\mathcal{I}$ is a set of states the option can be initiated into, $\pi$ is the policy that selects actions when the option is active, and $\beta$ is a random variable that gives the per-state probability of terminating the option. In general, learning options from scratch is hard, prone to collapse to single-action options, and less efficient than standard RL.
\citet{huang2019continuous} introduce Markov Jump Processes (MJPs), in which the agent both takes action and controls the frequency at which observations are received. Higher frequencies induce an increasing auxiliary cost to model scenarios where observations are limited. In essence, they propose to learn when to observe, while we propose to learn when to take control.
In a related way, \citet{biedenkapp2020towards} propose skip-MDPs, which decompose policies in the combination of a behavior policy (\textit{i.e.} which selects the action) and skip policy (\textit{i.e.} which selects the number of timesteps the action will be repeated for). 
Skip-MDPs does not exactly learn when to act as it permanently plays a decided action. This work rather shows that in most of situations an action need to be repeated in consecutive states, but as there are no states where the agent is deferring the control to an independent policy, they do not highlight the states where the agent decisions has a strong impact on the resulting behaviour and reward.
Note that dynamic action repetition~\citep{lakshminarayanan2017dynamic, sharma2017learning} is conceptually similar, but is not formalised as an abstraction over MDPs.

\textbf{Regularized RL.}
Regularization in RL~\citep{geist2019theory} is a well-studied topic. In particular, entropic regularization~\citep{neu2017unified} encourages learned policies to be as random as possible in all states. In contrast, when the default policy is uniform random, policies learned in lazy-MDPs are encouraged to be entirely random in a subset of all states only. Also, Kullback-Leibler regularization~\citep{vieillard2020leverage} encourages policies to stay close to their previous iterate during learning. In contrast, policies learned in lazy-MDPs act identically to the default policy in a subset of all states, and can act in arbitrary ways in the others.
Another way to ensure that the behavior of an agent does not diverge from a baseline behavior is to apply regularization on the state visitation distribution, instead of the action distribution induced by the policy ~\citep{lee2019efficient, geist2021concave}. 

\section{Lazy-MDPs}

\paragraph{MDP}
We use the Markov Decision Process (MDP) formalism~\citep{puterman1994mdp}. An MDP is a tuple $M = (\mathcal{S}, \mathcal{A}, \gamma, r, \mathcal{P}, \delta_0)$ where $\mathcal{S}$ is a state space, $\mathcal{A}$ is a discrete action space, $\gamma \in [0, 1]$ is a discount factor, $r \in [r_{min}, r_{max}]^{\mathcal{S} \times \mathcal{A}}$ is a reward function, $\mathcal{P} \in \Delta^{\mathcal{S} \times \mathcal{A}}_{\mathcal{S}}$ is a transition kernel (here $\Delta^X_Y$ is the set of functions that map an element of $X$ to a probability distribution over $Y$), and $\delta_0$ is the distribution of the initial state. We note the subspace of absorbing states $\mathcal{S}_{abs} \subseteq \mathcal{S}$. Absorbing states deterministically transition to themselves with zero rewards. In the following, we assume that we are in the infinite-horizon setting, and that $\gamma < 1$, but the proposed formalism is applicable to the finite-horizon setting as well.
Given an MDP, a policy $\pi\in \Delta^{\mathcal{A}}_{\mathcal{S}}$ maps states to probability distributions over actions is used to dict a behavior. The value function $V^{\pi}(s) = \mathbb{E}_{\pi}[\sum_{t=0}^{\infty}\gamma^t r(s_t,a_t) \vert s_0=s]$ measures the expectation of the delayed rewards by following a policy $\pi$ starting at $s$. Similarly, the action value function $Q^{\pi}(s,a) = \mathbb{E}_{\pi}[\sum_{t=0}^{\infty}\gamma^t r(s_t,a_t) \vert s_0=s, a_0=a]$ measures the expectation of the delayed rewards by following a policy $\pi$ starting with action $a$ at state $s$.
Another value function that we will use in section~\ref{eta_min_max} is $Z^{\pi}(s,a) = \mathbb{E}_\pi[\sum_{t=0}^{\infty}\gamma^t \mathbb{I}\{s_t \notin \mathcal{S}_{\text{abs}}\} \vert s_0=s, a_0=a\big]$, which is the expected discounted sum of steps before the agent meets a terminating state.

\paragraph{Lazy-MDP}
We now introduce the \textbf{Lazy Markov Decision Process} (\textit{lazy-MDP}). A lazy-MDP is a tuple $M_+ = (M, \Bar{a}, \bar{\pi}, \eta)$, where $M = (\mathcal{S}, \mathcal{A}, \gamma, r, \mathcal{P}, \delta_0)$ is the \textit{base} MDP, $\bar{a}$ the lazy action that defers decision making to the default policy $\bar{\pi} \in \Delta^{\mathcal{S}}_{\mathcal{A}}$ and $\eta \in \mathbb{R}$ is a penalty.
The reward function is that of the base MDP, except that all actions but the lazy one incur an additional reward of $-\eta$.
Hence, a lazy-MDP is also an MDP. It can be written as $M_+ = (\mathcal{S}, \mathcal{A_+}, \gamma, r_+, \mathcal{P}_+, \delta_0)$. While $\mathcal{S}, \gamma, \delta_0$ are conserved from the base MDP, $\mathcal{A_+}, r_+$ and $\mathcal{P}_+$ depend on their equivalents in the base MDP, and on $\bar{\pi}$ and $\eta$:
\begin{equation}
    \mathcal{A}_+ = \mathcal{A} \cup \{ \bar{a} \},
\end{equation}
\begin{equation}
    r_+(s, a) = 
    \begin{cases}
    r(s, a) - \eta, & \text{if } a \in \mathcal{A},\\
    \sum_{a\in\mathcal{A}} \bar{\pi}(a|s) r(s, a), & \text{if } a = \bar{a},
    \end{cases}
\end{equation}
\begin{equation}
    \mathcal{P}_+(s' | s, a) =
    \begin{cases}
    \mathcal{P}(s' | s, a), & \text{if } a \in \mathcal{A},\\
    \sum_{a\in\mathcal{A}} \bar{\pi}(a|s) \mathcal{P}(s'|s,a), & \text{if } a = \bar{a}. 
    \end{cases}
\end{equation}
In what follows, we will use the notation $X_{+}$ to distinguish functions or distributions over the augmented action space $\mathcal{A}_{+}$.

\section{Optimality in lazy-MDPs}

In this section, we provide a characterization of the optimality in lazy-MDPs, similarly to what is done for regular MDPs.
The derived results have two main implications. First (in Sec.~\ref{4.1} and \ref{4.2}), we identify what we call the \emph{lazy-gap}, which quantifies the importance of taking control or not in a given state. Second (in Sec.~\ref{4.3}), we show that taking control or not depending on the sole value of this lazy-gap leads to optimal behavior in the lazy-MDP.
All statements are proven in the Appendix. 

\subsection{Value functions}\label{4.0}

Let $\pi_+(a\in\mathcal{A}_+\vert s)$ be a policy in the lazy-MDP. If the agent chooses the lazy action $\bar{a}$, the performed action $a\in\mathcal{A}$ is sampled according to the default policy $\bar{\pi}$. We formalize the resulting \textit{lazy policy} (in the base MDP) as follows:
\begin{align}\label{pi_mdp}
    \pi(a\in\mathcal{A}\vert s) &= P\Big[(a\sim \pi_+) \cup (\bar{a} \sim \pi_+ \cap a\sim \bar{\pi})\Big], \\
    &= \pi_+(a\vert s) + \pi_+(\bar{a} \vert s) \bar{\pi}(a\vert s),
\end{align}
\looseness=-1
satisfying $\sum_{a\in\mathcal{A}} \pi(a \vert s) = 1$. A crucial point is that $\pi$ has the same dynamics in the base MDP as $\pi_+$ in the corresponding lazy-MDP. We are interested in the value function $V_+^{\pi_+}(s)$, which is the value of $\pi_+$ in the lazy-MDP, and takes the penalties into account. We would like to decompose $V_+^{\pi_+}(s)$ as a function of $V^\pi(s)$ (\textit{i.e.} the value function associated with $\pi$ in the base MDP) and a cost function $C^{\pi_+}(s)$:
\begin{equation}
    V_+^{\pi_+}(s) = V^\pi(s) + C^{\pi_+}(s).
\end{equation}
\begin{theorem}\label{th1}
    $C^{\pi_+}(s)$ satisfies the following Bellman equation:
    \begin{equation}
        C^{\pi_+}(s) = -\eta(1- \pi_+(\bar{a}\vert s)) + \gamma \mathbb{E}_{a\sim \pi, s'\sim \mathcal{P}(\cdot \vert s,a)} C^{\pi_+}(s'). 
    \end{equation}
\end{theorem}

 While $V^\pi$ is the expected discounted sum of rewards obtained by following $\pi$ in the base MDP, the cost $C^{\pi_+}$ can be interpreted as the expected discounted sum of the incurred penalties. For instance, a policy that never picks the lazy action (\textit{i.e.} $\forall s, \; \pi_+(\bar{a}\vert s)=0$) gets a maximal cost:
\begin{equation}
    V_+^{\pi_+} = V^\pi - \frac{\eta}{1-\gamma}.
\end{equation}

\subsection{Q-functions}\label{4.1}

Let $Q_{\setminus \bar{a}}^{\pi_+}(s, a\in\mathcal{A})$ be the value (in the lazy-MDP) of taking another action than the default one:
\begin{align}
    Q_{\setminus \bar{a}}^{\pi_+}(s,a) &= r(s,a) - \eta +\gamma \mathbb{E}_{s'}\Big[V_+^{\pi_+}(s')\Big]\\
    &= r(s,a) - \eta +\gamma \mathbb{E}_{s'}\Big[V^\pi(s') + C^{\pi_+}(s')\Big]\\
    &= Q^\pi(s,a) -\eta + \gamma\mathbb{E}_{s'}\Big[C^{\pi_+}(s')\Big],
\end{align}
and $\pi_{\setminus \bar{a}}(a\in\mathcal{A}\vert s)$ be the policy obtained by excluding the lazy action from $\pi_+$, \textit{i.e.} assuming $\pi_+(\bar{a}\vert s)<1$:
\begin{equation}
    \forall a \in \mathcal{A},\hspace{2mm}\pi_{\setminus \bar{a}}(a\vert s) = \frac{\pi_+(a\vert s)}{\sum\limits_{a'\neq\bar{a}}\pi_+(a'\vert s)} = \frac{\pi_+(a\vert s)}{1-\pi_+(\bar{a}\vert s)}.
\end{equation}
We can express the value function $V_+^{\pi_+}$ as a function of $Q_{\setminus \bar{a}}^{\pi_+}$:
\begin{prop}\label{p1}
\begin{align}
    V_+^{\pi_+}(s) =& (1 - \pi_+(\bar{a}\vert s)) \mathbb{E}_{a\sim\pi_{\setminus \bar{a}}}\Big[Q_{\setminus \bar{a}}^{\pi_+}(s,a)\Big]\\ 
    &+ \pi_+(\bar{a}\vert s) \bigg(\mathbb{E}_{a\sim\bar{\pi}}\Big[Q_{\setminus \bar{a}}^{\pi_+}(s,a)\Big] + \eta\bigg).
\end{align}
\end{prop}
We then have the expression for $Q^{\pi_+}_+(s,a\in\mathcal{A}_+)$, the Q-function of $\pi_+$ in the lazy-MDP:
\begin{prop}\label{p2}
\begin{align}
    Q^{\pi_+}_+(s,a) &=
    \begin{cases}
        Q_{\setminus \bar{a}}^{\pi_+}(s,a) &\text{if } a\neq\bar{a},\\
        \mathbb{E}_{a\sim\bar{\pi}}\Big[Q_{\setminus \bar{a}}^{\pi_+}(s,a)\Big] +\eta &\text{if } a=\bar{a}.
    \end{cases}
\end{align}
\end{prop}

\subsection{Greediness}\label{4.2}

A policy $\pi_+$ is greedy wrt 
$Q_+$ (noted $\pi_+\in \mathcal{G}(Q_+)$) if and only if:
\begin{equation}
    \forall s \in \mathcal{S}, \hspace{3mm} \pi_+(\cdot \vert s) \in \argmax\limits_{\pi_+(\cdot \vert s)} \mathbb{E}_{a\sim\pi_+}\Big[Q_+(s,a)\Big].
\end{equation}
Given a Q-function $Q$, a useful quantity to construct a greedy policy is what we call the \textbf{\textit{lazy-gap}}, noted $G_Q(s)$:
\begin{equation}
    G_Q(s) = \max_{\mathcal{A}}Q(s,\cdot) - \mathbb{E}_{\bar{\pi}}\Big[Q(s,a)\Big],
\end{equation}
which is the gap between the value of the best action (in $\mathcal{A}$) and the expected action-value when the immediate next action is picked by the default policy (as defined by $Q$) \textit{in the lazy-MDP} (\textit{i.e.}, with costs taken into account). 
In order to be greedy with respect to a policy $Q_+$, one needs to choose the lazy action if $G_{Q_{+\setminus\bar{a}}}(s)\leq\eta$, and to take the argmax of $Q_{+\setminus\bar{a}}$ otherwise (over $\mathcal{A}$).
\begin{prop}\label{p3}
The following policy $\pi_+$ is greedy with respect to $Q_+$:
\begin{align}
    \pi_+(\cdot \vert s) &= 
    \begin{cases}
    \frac{\mathbb{I}\big\{a\in\argmax_\mathcal{A} Q_{+\setminus\bar{a}}(s,a)\big\}}{\big\lvert \argmax_\mathcal{A} Q_{+\setminus\bar{a}}(s,a) \big\rvert} &\text{if } G_{Q_{+\setminus\bar{a}}}(s)>\eta,\\
    \mathbb{I}\Big\{a=\bar{a}\Big\} &\text{otherwise}.
    \end{cases}
\end{align}
\end{prop}
\looseness=-1
Since the greediness as constructed above does not depend on the value of the lazy action $Q_+(s, \bar{a})$, we can define a greedy policy in the lazy-MDP with respect to a Q-function of the base MDP:
\begin{align}
    \mathcal{G}(Q)(\cdot \vert s) &:= 
    \begin{cases}
    \frac{\mathbb{I}\big\{a\in\argmax_\mathcal{A} Q(s,a)\big\}}{\big\lvert \argmax_\mathcal{A} Q(s,a) \big\rvert} &\text{if } G_{Q}(s)>\eta,\\
    \mathbb{I}\Big\{a=\bar{a}\Big\} &\text{otherwise}.
    \end{cases}
\end{align}

\subsection{Optimality}\label{4.3}

We define the greedy operator $\mathcal{T}$, that maps a Q-function to the immediate reward plus the average value of the next state according to the greedy policy:
\begin{equation}
    \mathcal{T}Q(s, a \in \mathcal{A}) \eqdef
    r(s,a) -\eta + 
    \gamma\mathbb{E}_{s'\sim \mathcal{P}(\cdot \vert s,a), a'\sim \mathcal{G}(Q)(\cdot \vert s')}\Big[Q(s',a')\Big].
\end{equation}
\begin{theorem}\label{th:convergence}
    $\mathcal{T}$ is a $\gamma$-contraction, and converges to $Q^*:=Q_{\setminus \bar{a}}^{\pi^*_+}$ where $\pi^*_+$ is the optimal policy in the lazy-MDP.
\end{theorem}

 This allows us to identify the optimal policy to take decisions in the augmented action space $\mathcal{A}_+$.
\begin{cor}
    \label{cor:optimality}
    $\pi^*_+$ is a deterministic policy that verifies $\pi^*_+(\Bar{a} | s)>0 $ if and only if $G^*(s) > \eta$, with $G^* = G_{Q^*}$ is the lazy-gap under the optimal action-value in the lazy-MDP.
\end{cor}

\section{Setting the cost of taking control}\label{eta_min_max}

Given a known base MDP, we may want to find the minimal cost $\eta_{\min}$ such that an optimal policy takes the lazy action in at least one state, as well as the maximal cost $\eta_{\max}$ such that an optimal policy does not take the lazy action in all states. 
From Prop.~\ref{cor:optimality}:
\begin{align}
    \eta_{\min} &= \inf\Big\{\eta>0 \text{ s.t. } \exists s,\hspace{1mm} G^*(s) < \eta \Big\},\\
    \eta_{\max} &= \sup\Big\{\eta>0 \text{ s.t. } \exists s,\hspace{1mm} G^*(s) > \eta \Big\},
\end{align}
\looseness=-1
where $G^*$ is the lazy-gap associated with the optimal value function $Q_{\setminus \bar{a}}^{\pi^*_+}$. Taking $\eta$ between these two values allows to train agents that decide when to act in a non-trivial way. One can then equate states where the agent takes control as important states, under the right $\eta$.

\subsection{\boldmath\texorpdfstring{$\eta_{\max}$}{eta max}}

When $\eta$ is equal or larger than the lazy-gap in all states, the optimal policy consists in deferring all actions to the default policy, which induces no cost. Thus, $\eta_{\max}$ is simply equal to the maximal lazy-gap under the default policy.
\begin{theorem}\label{th3}
Let $Q^{\bar{\pi}}(s, a\in \mathcal{A})$ be the Q-function of the default policy in the base MDP. Then: $\eta_\text{max} = \max_s G_{Q^{\bar{\pi}}}(s)$.
\end{theorem}

\subsection{\boldmath\texorpdfstring{$\eta_{\min}$}{eta min}}

One cannot apply a similar treatment to $\eta_{\min}$, due to most actions being non-default and corresponding costs having to be taken into account. However, if $\eta$ is smaller or equal to the lazy-gap in all states, an optimal agent will follow the optimal policy of the base MDP $\pi^*$, and the incurred cost will be equal to $-\eta$ multiplied by the fictitious Q-value associated with $\pi^*$ for a reward function that is 0 in absorbing states and 1 otherwise:
\begin{equation}
    Z^{\pi^*}(s,a) = \mathbb{I}\{s \notin \mathcal{S}_{abs}\} + \gamma \mathbb{E}_{s'}\bigg[ \mathbb{E}_{\pi^*}\Big[Z^{\pi^*}(s', \cdot)\Big]\bigg]
\end{equation}
By construction $C^{\pi^*}(s) = -\eta \mathbb{E}_{\pi^*}[Z^{\pi^*}(s, \cdot)]$, where $C^{\pi^*}(s)$ is the cost for always following $\pi^*$ from $s$. If no absorbing state is ever reached, we have $Z^{\pi^*}(s,a) = \frac{1}{1-\gamma}$ for all state $s$ and action $a$. However, in practice the MDPs we consider have terminal states and eventually end. $Z^{\pi^*}$ can thus have different values under different state-action couples, impacting the value of $\eta_\text{min}$. Its value is given by the following theorem:

\begin{theorem}\label{th:eta_min}
Let $\pi^*$ be the optimal policy in the base MDP, and $Q^*(s, a\in \mathcal{A})$ the associated Q-function. Then:
\begin{equation}
    \eta_\text{min} =
     \min_s\max_a \frac{Q^*(s,a) - \mathbb{E}_{\bar{\pi}}\Big[Q^*(s, \cdot)\Big]}{ 1 + \bigg(\mathbb{E}_{\pi^*}\Big[Z^{\pi^*}(s, \cdot)\Big] - \mathbb{E}_{\bar{\pi}}\Big[Z^{\pi^*}(s, \cdot)\Big]\bigg)},
\end{equation}
with $\eta_\text{min} \geq 0$.
\end{theorem}
We empirically validate these boundaries over different lazy-MDPs and report the results in Appendix~\ref{appendix:eta_min_max}. As expected, when $\eta < \eta_{\min}$ no lazy actions are ever selected, and when $\eta > \eta_{\max}$, the agent always chooses the lazy action. 
We discuss how to approximate $\eta_\text{min}$ and $\eta_\text{max}$ in the next section.

\section{Learning when and how to act in Lazy-MDPs}

Since lazy-MDPs can be described as augmented MDPs, standard RL algorithms can still be used to provide policies that maximize the cumulative sum of rewards (which include a cost when taking control). As a result, by converting an MDP into a lazy-MDP, RL agents learn when and how to act without any change to their workings. We now discuss design choices for the two parameters of lazy-MDPs: the cost $\eta$ and the default policy $\bar{\pi}$. 

\textbf{Value of cost.} 
Regarding $\eta$, the explicit expressions for the bounds we provided guarantee meaningful behavior from optimal policies (\textit{i.e.} not always defaulting and not always taking control). Estimating $\eta_\text{max}$ is feasible since the default policy $\bar{\pi}$ is supposed available. It requires to estimate its action-values (for instance, using SARSA~\citep{rummery1994online}) so that the maximal lazy-gap can be approximated (either by taking the maximum gap across the known set of states, or using rollouts to get approximate coverage). To estimate $\eta_\text{min}$, usually one does not have access to the optimal Q-function $Q^*$ of the base MDP nor to $Z^{\pi^*}$. In that case, a solution is to use value iteration or Q-learning~\citep{watkins1992q} to get approximations $Q_\theta$ and $Z_\theta$, where $Z_\theta$ is obtained by replacing all the rewards by ones in the loss used to learn $Q_\theta$. %

\textbf{Choice of the default policy.} 
Regarding the choice of $\bar{\pi}$, we argue that taking a random policy (with, say, uniform action probabilities) is the simplest option available: it does not require any knowledge about the task, and is on par with the idea that the agent should take control only when actions actually matter (\textit{i.e.}, a specific action is noticeably better than uniform sampling). Doing so results in a type of regularization that is conceptually close to entropic regularization, except that the agent has incentive to be as random as possible \textit{in a subset of the states only}. In some cases, including complex scenarios, alternative options might be preferable: having to take control too often could lead to a high cumulative cost and discourage exploration, unless $\eta$ is properly tuned. Similarly to residual learning, an interesting substitute is a known, suboptimal policy. In that case, the agent has incentive to take control in states where the base policy is noticeably suboptimal. %

\textbf{Interpretability of lazy policies.} 
Due to the cost of taking control, the lazy policies that are learned should only take control in a handful of states. We hope that this leads to increased interpretability for several reasons: the subset of states (and the corresponding controls) can be assessed against expert knowledge, and the performance of the learned policy can be compared to that of the base policy to ensure that the gains are sufficient and justify the overhead. 
Specifically, we argue that there are two special cases where lazy actions indeed give information about the \textit{overall} importance (and unimportance) of states:
\begin{enumerate}
    \item with a uniform random policy as default (and the right penalty), an optimal agent should only pick its actions when selecting among optimal actions brings a substantial advantage over picking the action at random. Therefore, we argue that states where the agent defers its actions are likely to be unimportant in the sense that the agent is content with acting randomly.
    \item with a mixture between optimal and uniform random policy as default (and the right penalty), an optimal agent should only pick its actions when selecting among optimal actions brings a substantial advantage over \textit{often} selecting among optimal actions. Therefore, we argue that states where the agent picks its actions are likely to be important in the sense that the agent is \textit{not} content with acting \textit{almost} optimally.
\end{enumerate}
More broadly speaking, lazy actions give information about \textit{the specific importance of states so as to improve on the performance of the default policy}. 

We study the interpretability of solutions empirically in the next section. 

\section{Experiments}

In this section, we empirically address the following questions: \textbf{1)}~Do policies from lazy-MDPs learn to take control when it matters? \textbf{2)}~Are the partitions of states where the agent decides or not to act interpretable? \textbf{3)}~How does reducing the frequency of agent controls (by increasing the cost $\eta$) affect its returns? Details about implementations and the choice of hyperparameters can be found in Appendix~\ref{implementation_details}.

\subsection{Taking control when it matters}
We first study the behavior of lazy policies on small discrete problems, where the exact solutions can be approached as well as values for $\eta_\text{min}$ and $\eta_\text{max}$.

\begin{figure}[t]
    \centering
    \begin{subfigure}[b]{0.25\linewidth}
        \centering
        \includegraphics[width=\linewidth]{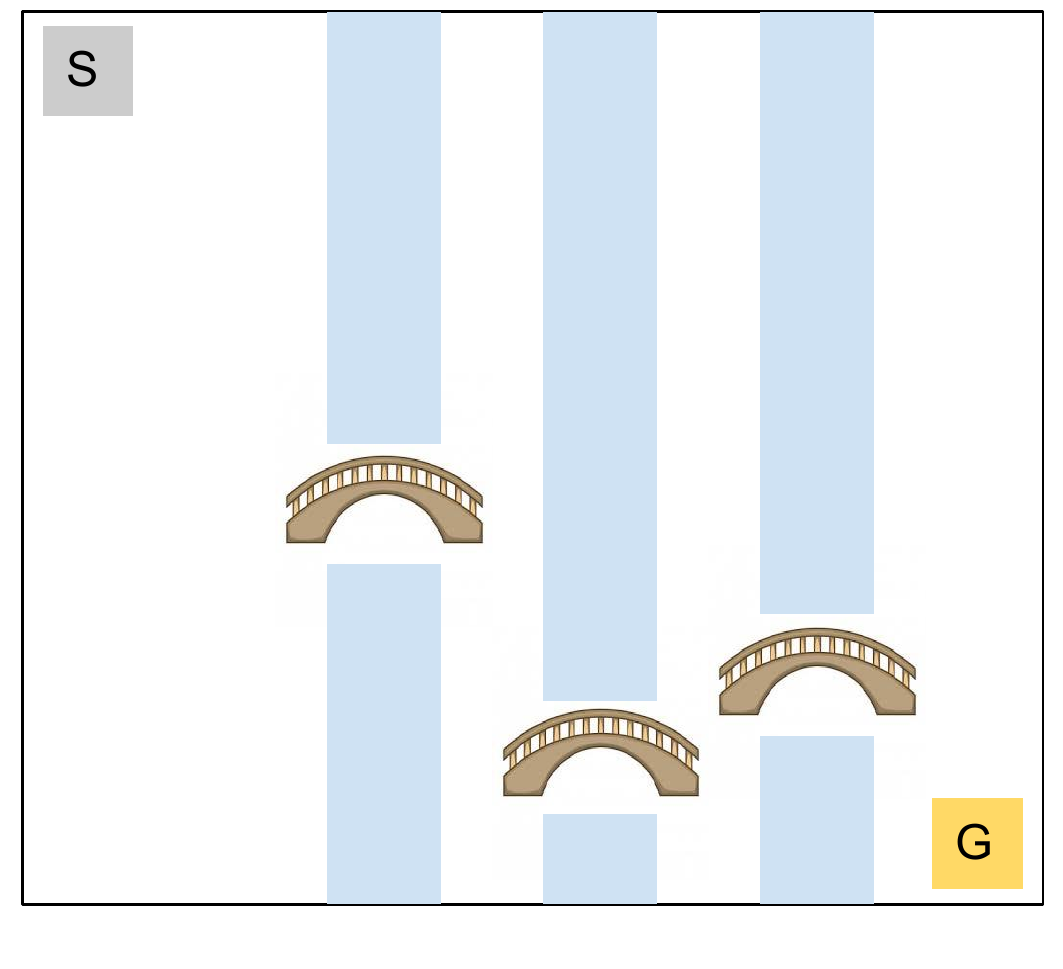}
    \end{subfigure}
    \begin{subfigure}[b]{0.225\linewidth}
        \centering
        \includegraphics[width=\linewidth]{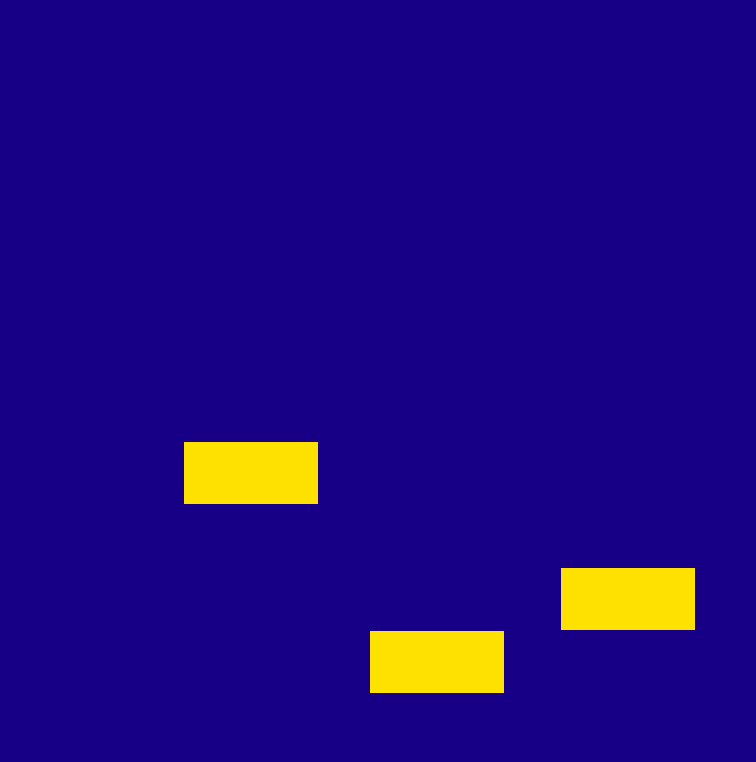}
    \end{subfigure}
    \caption{\textbf{Left:} Rivers and Bridges environment. The agent starts up left (S) and has to cross the rivers through the bridges to reach the goal point (G). Falling in the water is punished by $\mathcal{R}=-100$ and reaching the goal is rewarded by $\mathcal{R}=1$. The default policy is the optimal $\pi^*$ everywhere but on the bridges where it is uniformly random. \textbf{Right:} Heatmap of the resulting lazy-gap using $\eta=\eta_{\min}=0$. As expected the lazy-gap is zero everywhere but on the bridges where optimal agents learn to take control. This result is valid for any value of $\eta < \eta_{\max}$.}
    \label{fig:rivers}
\end{figure}

\textbf{Rivers and Bridges.}
A simple environment that illustrates how lazy-MDPs work is a gridworld involving some dangerous pathways -- where the agent needs to provide precise controls, while other states are safe -- the agent can rely on the default policy. We implemented a basic scenario in which the agent has to cross three rivers by taking slippery bridges. We call this environment Rivers and Bridges (R\&B), which is illustrated in Fig~\ref{fig:rivers}. Falling in the water is penalized by a strong negative reward ($\mathcal{R}=-100$), while reaching the goal point beyond the rivers results in a small positive reward ($\mathcal{R}=1$). To simulate the slipperiness of the bridges, we take as default policy  the policy that is optimal everywhere but on the bridges where it is uniformly random. That way, the agent should trust the default policy everywhere but on the bridges where it should take the control despite the cost. As there is at least one state where the policy is optimal, applying Theorem~\ref{th:eta_min}, we get $\eta_{\min}=0$. As shown in Fig.~\ref{fig:rivers} right, we verify that for any $\eta$ such that $0 < \eta \leq \eta_{\max}$, the lazy-gap $G^*(s)$ is only positive on the bridges, which means an optimal agent only takes control in those states, and justifies the application of lazy-MDPs to evaluate where and when to trust a default behavior. 

\begin{figure}[t!]
    \centering
    \begin{subfigure}[b]{0.25\linewidth}
        \centering
        \includegraphics[width=\linewidth]{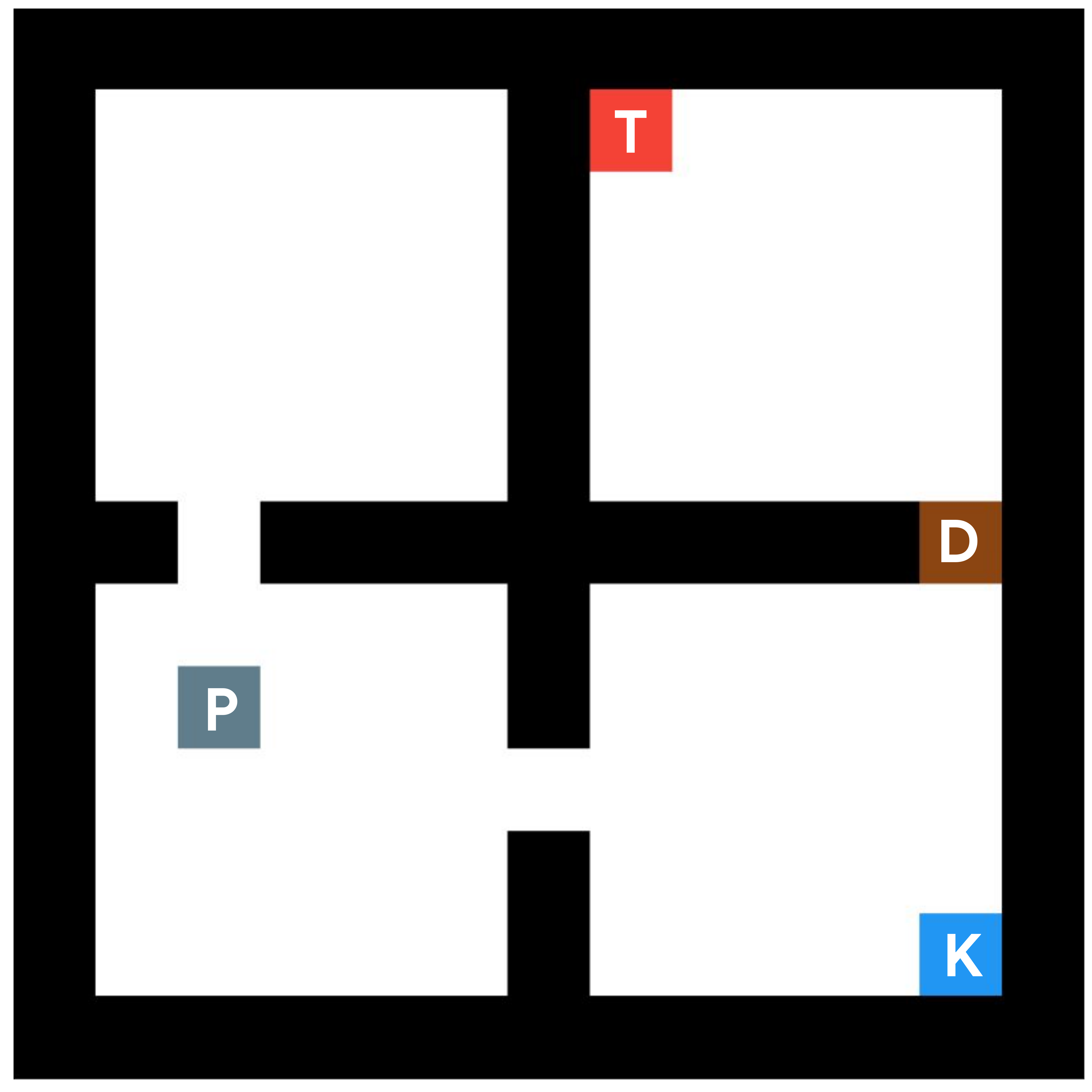}
        \caption{}
      \end{subfigure}%
      \begin{subfigure}[b]{0.25\linewidth}
        \centering
        \includegraphics[width=\linewidth]{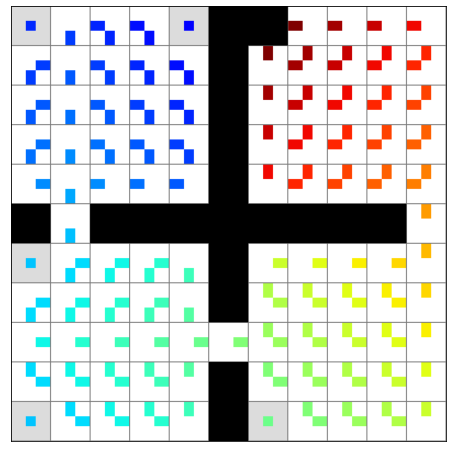} 
        \caption{$\eta = 0.008$}
      \end{subfigure}%
      \begin{subfigure}[b]{0.25\linewidth}
        \centering
        \includegraphics[width=\linewidth]{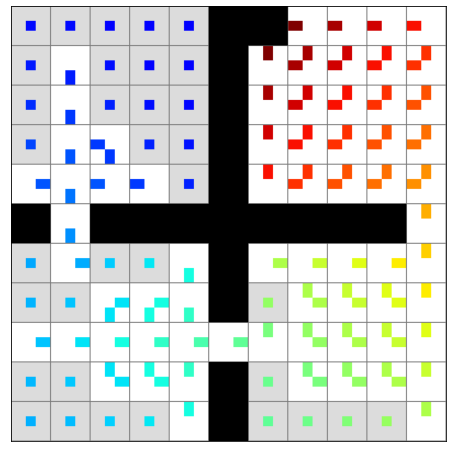}
        \caption{$\eta = 0.02$}
      \end{subfigure}%
      \begin{subfigure}[b]{0.25\linewidth}
        \centering
        \includegraphics[width=\linewidth]{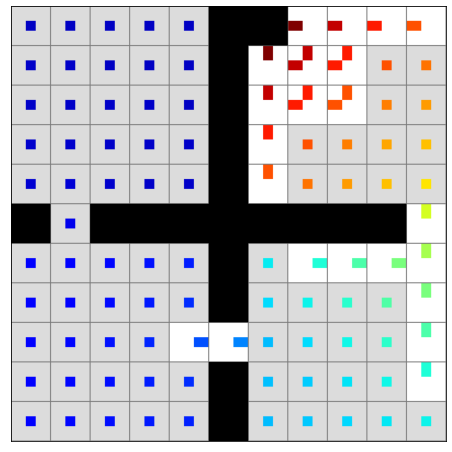}
        \caption{$\eta = 0.05$}
      \end{subfigure} 
    \caption{\textbf{(a):} Key-Door-Treasure environment. \textbf{(b-d):} Policies learned in the lazy-MDP by value iteration, with a uniform random default policy, and increasing $\eta$. Centered points in gray cells identify lazy actions, color indicates learned action-values (blue=lowest, red=highest). With $\eta$ increasing, the policy learns to reduce its controls to states that allow it to progress from one room to another as well as those in the vicinity of the goal state.}
    \label{fig:kdt}
\end{figure}

\textbf{Key-Door-Treasure}~\citep{oh2018self} (KDT) is a variant of the classic Four Rooms task~\citep{sutton1998reinforcement} with a harder exploration problem: the agent needs to grab a key, open the door and get to the location of the treasure (in that order) to solve the task. The agent is only rewarded when reaching the treasure. We study the nature of the solutions learned in the lazy-MDP version of KDT under several values of $\eta$, with a uniform random default policy. The results are shown in Fig.~\ref{fig:kdt}. They match our intuition: the higher the value of $\eta$, the fewer the states in which the agent takes control; until the agent only acts in the most crucial states (\textit{i.e.} to pass from one room to another or to get to the treasure).  

\subsection{Interpretability}
\label{subsection:interpretability}

\begin{figure*}[t]
    \centering
    \includegraphics[width=\textwidth]{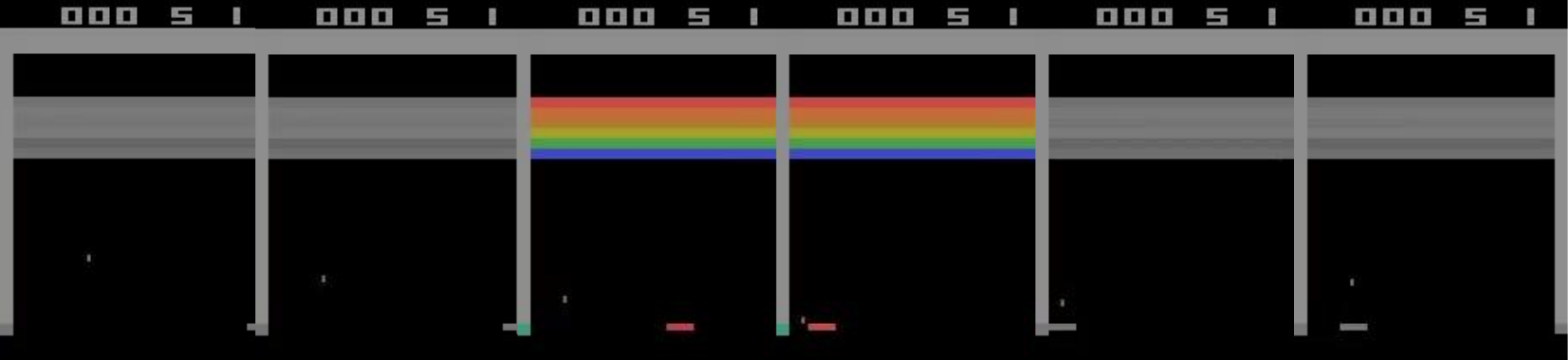}
    \caption{Illustration of the policy learned in the lazy-MDP version of Breakout ($\eta=0.1$, with a uniform random default policy): the agent learns to take control (\textbf{colored frames}) only moments before the ball has to be hit in order not to lose.}
    \label{fig:breakout}
\end{figure*}

\begin{figure*}[ht!]
    \centering
    \includegraphics[width=\textwidth]{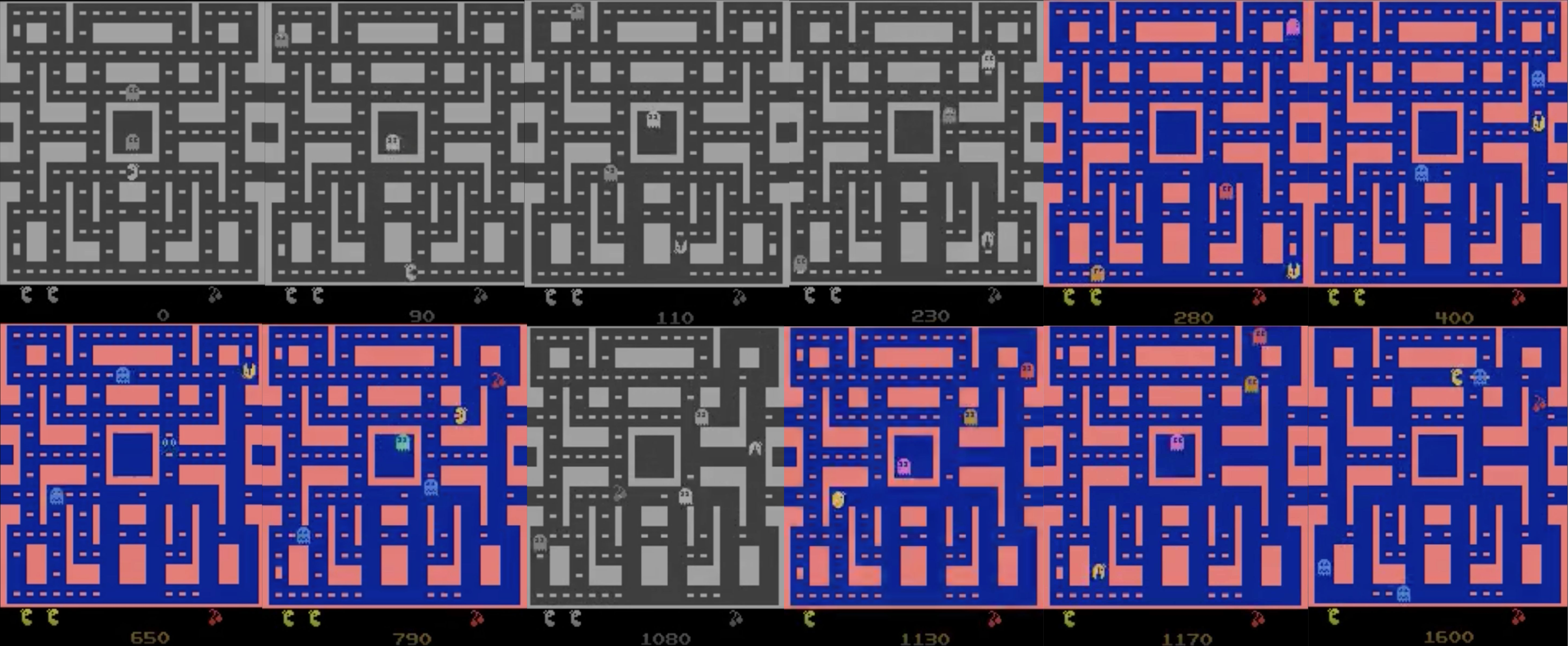}
    \caption{Illustration of the policy learned in the lazy-MDP version of Ms Pacman ($\eta=0.5$, with a uniform random default policy). 
    Frames are ordered from left to right, top to bottom, and correspond to non-consecutive frames from a single episode of interaction.
    At the beginning of the episode, the agent is immobile for a few frames, during which it defaults its actions to avoid penalties (\textbf{frame 1}). Once free to move, it keeps defaulting its actions and collecting nearby bonuses (\textbf{frames 2-4}) until ghosts become close enough. The agent then sparsely takes control (colored frames), be it to obtain power-ups (\textbf{frames 5 \& 7}), and later on eat the ghosts (\textbf{frames 6 \& 12}), or collect several bonuses in a row (\textbf{frames 10-11}).}
    \label{fig:mspacman}
\end{figure*}

\begin{figure*}[ht!]
    \centering
    \includegraphics[width=\textwidth]{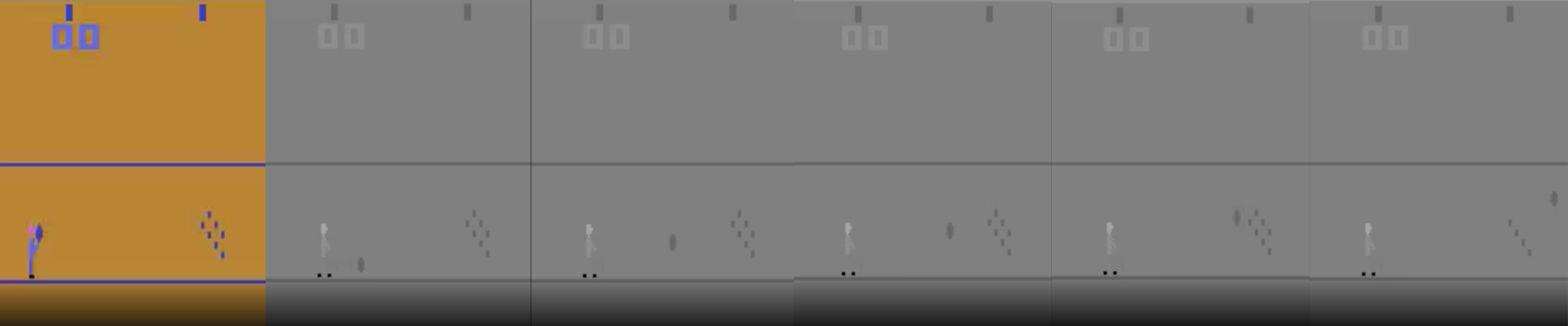}
    \caption{Illustration of the policy learned in the lazy-MDP version of Bowling ($\eta=0.04$, with a uniform random default policy). 
    The agent learns to only take control \textbf{(colored frame}) when preparing to throw the ball. This is the only time it can control the orientation of the throw, which is key to knock pins down. In subsequent timesteps, its actions have no effects on the ball, and accordingly the lazy action is picked instead.}
    \label{fig:bowling}
\end{figure*}

To study interpretability, we use more complex environments where the behaviour of an RL agents is not trivially interpretable. We make the hypothesis that lazy-MDPs can help at explaining which states and what actions are important in order to get high returns. In this study, we opt for a qualitative measure of importance and show (via corresponding frames) the states of importance as the ones where the agent decides to take control over the uniform random, default policy. We look at lazy policies learned in the lazy-MDP version of Atari 2600 games from the Arcade Learning Environment~\citep{bellemare2013arcade}. We focus on games where timing plays a central role, such as Pong, Breakout and Ms Pacman. We use a standard DQN agent~\citep{mnih2015humanlevel}, whose implementation we take from the Dopamine framework~\citep{castro2018dopamine}. We display a representative portion of a lazy agent trajectory in Breakout in Fig.~\ref{fig:breakout}, which is well aligned with our intuition of the timing of this task: critical controls happen when the ball gets back to the paddle, while the remaining controls have limited impact on subsequent success. We also display key moments of a lazy agent trajectory in Ms Pacman in Fig.~\ref{fig:mspacman}. The agent alternates between defaulting (most of the time) and taking control (sparsely, either for a single frame or a sequence of frames) in order to escape ghosts, to obtain power-ups and defeat ghosts, or to collect multiple bonuses in a row. Finally, we display a representative portion of episode in Bowling in Fig.~\ref{fig:bowling}. The agent takes control when aiming with the ball, and defers control to the default policy when the ball is moving towards the pins, during which actions have no effects on the outcome of the throw.
All in all, the timing of the agent controls matches our intuitions.

We also compared the importance as quantified by the lazy-gap with usual measures, such as the action-gap~\citep{bellemare2016increasing}, which measures the difference between the best and the second best action values at a given state ($\max_{\mathcal{A}}Q(s, \cdot) - \max_{\mathcal{A}\setminus a^*} Q(s, \cdot)$), and importance advice~\citep{torrey2013teaching}, which measures the difference between the best and the worst action values at a given state ($\max_{\mathcal{A}}Q(s, \cdot) - \min_{\mathcal{A}}Q(s, \cdot)$). Fig.~\ref{fig:importance} in Appendix~\ref{annex:importance} displays the state importance according to these measures on the KDT environment. As visible, the lazy-gap only attributes importance to states with key actions (picking up the key, passing through doors, reaching the treasure). On the other hand, the action-gap uniformly emphasizes all states along the trajectory of the optimal policy, while the importance advice is dominated by the proximity to the reward and does not discriminate key actions.

\subsection{Lazy exploration}
\label{subsec:exploration}

\begin{figure}[t!]
    \centering
    \begin{subfigure}[b]{0.22\linewidth}
        \centering
        \includegraphics[width=\linewidth]{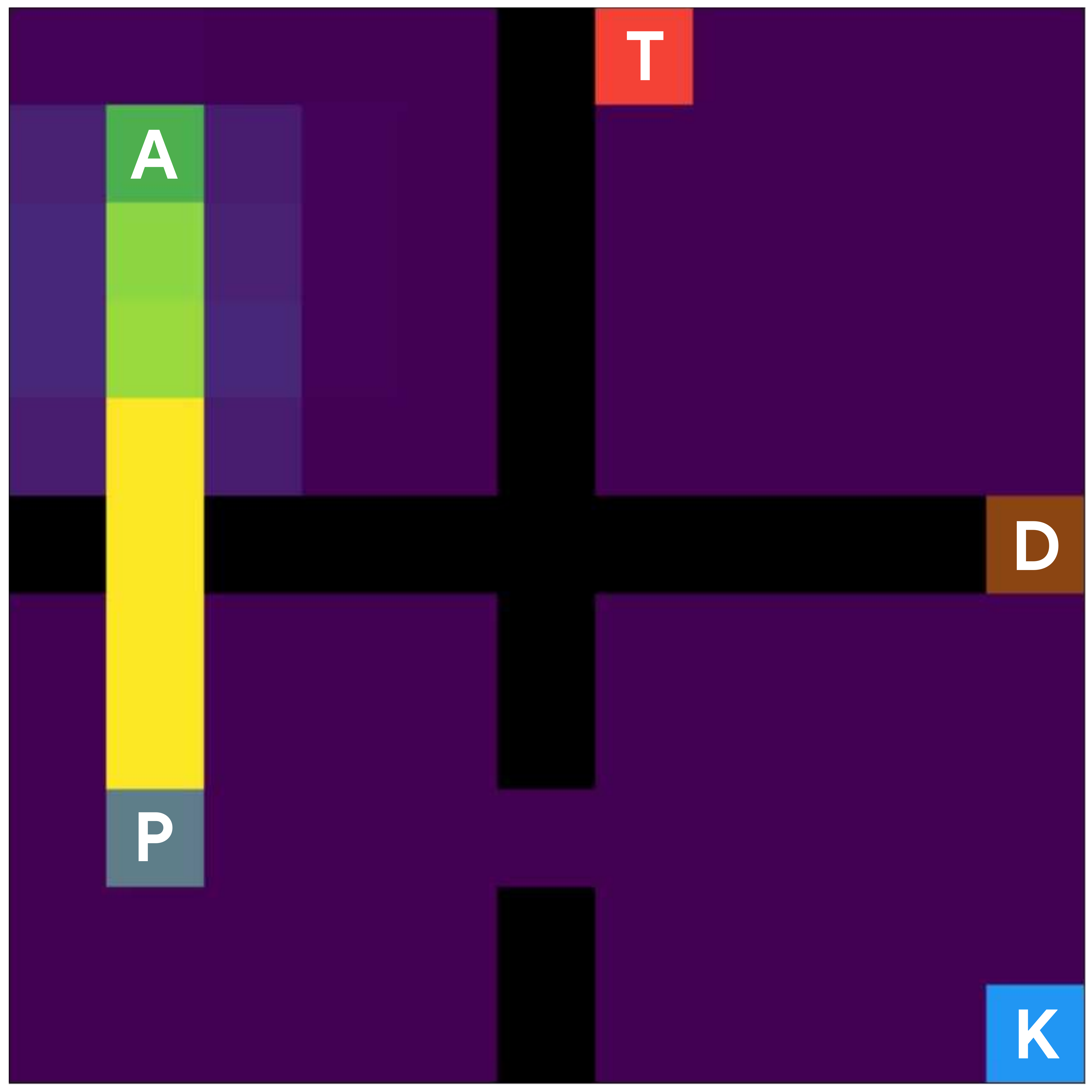}
        \caption{$\eta = 0$}
      \end{subfigure}%
      \begin{subfigure}[b]{0.22\linewidth}
        \centering
        \includegraphics[width=\linewidth]{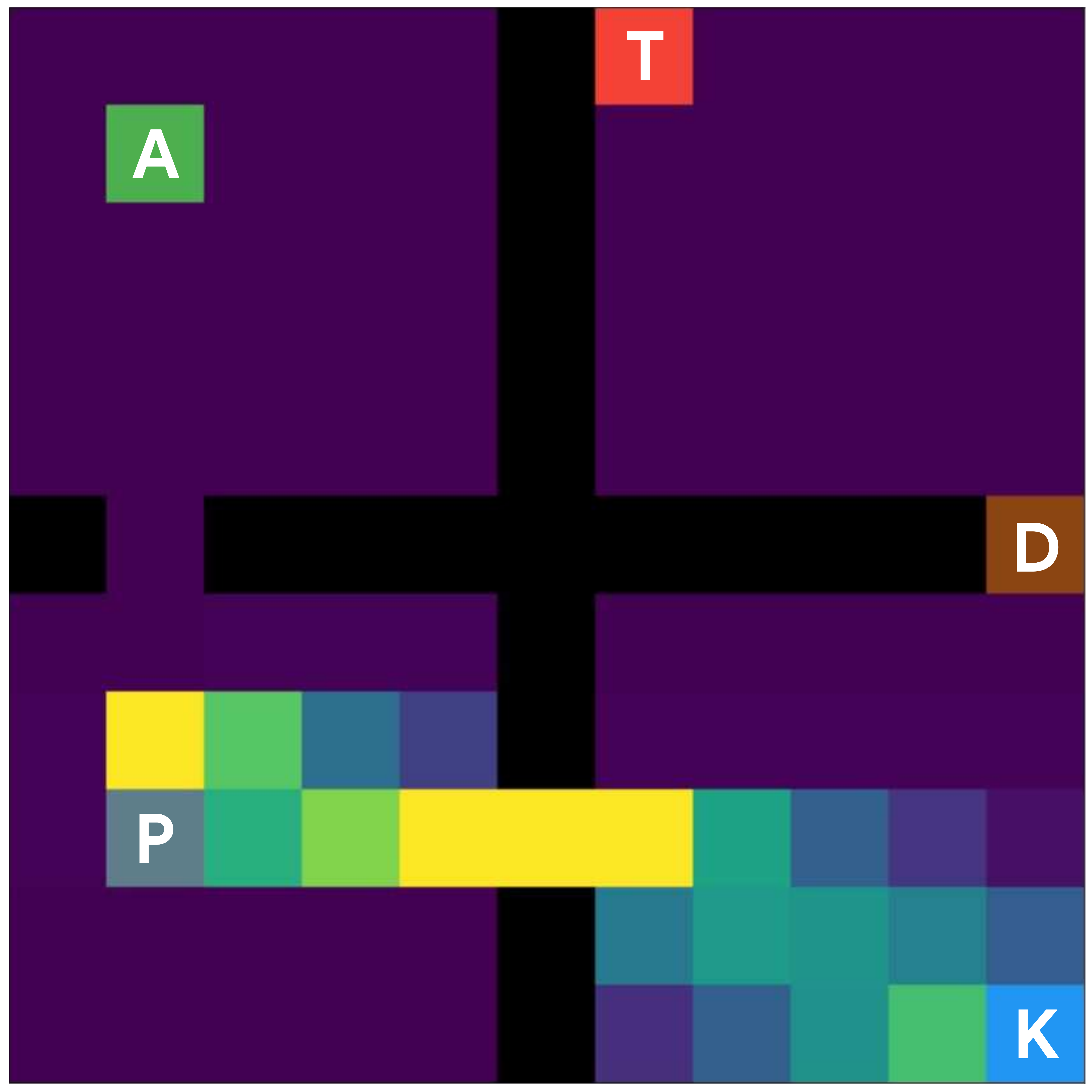} 
        \caption{$\eta = 0.03$}
      \end{subfigure}%
      \begin{subfigure}[b]{0.22\linewidth}
        \centering
        \includegraphics[width=\linewidth]{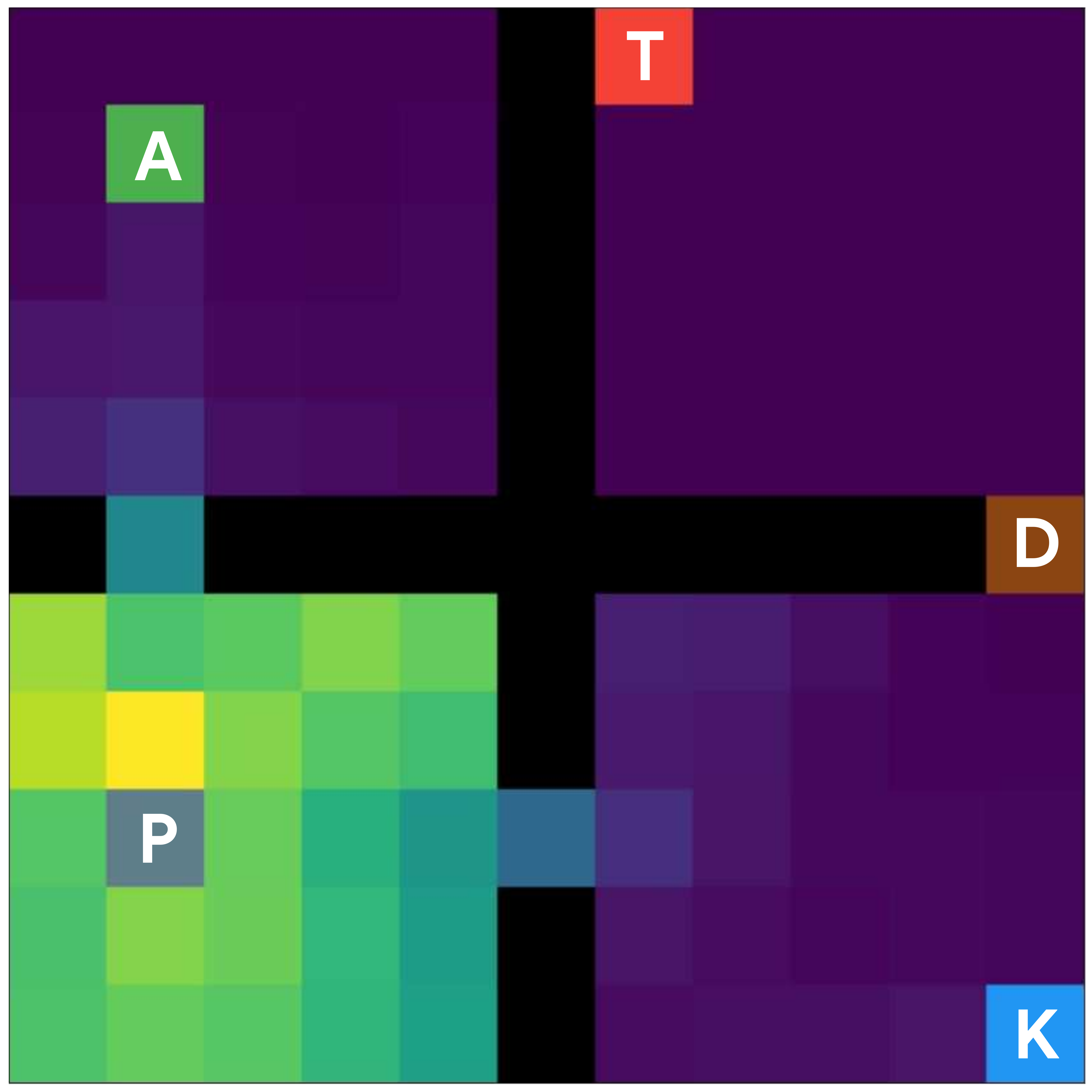}
        \caption{$\eta = 0.05$}
      \end{subfigure}%
      \begin{subfigure}[b]{0.34\linewidth}
        \centering
        \includegraphics[width=\linewidth]{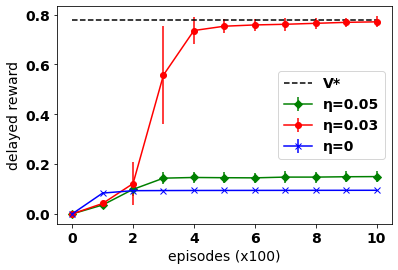}
        \caption{}
      \end{subfigure} 
    \caption{We add an apple which grants a small reward to KDT. It acts as a distractor for exploration. \textbf{(a-c):} Asymptotic state occupancy measure of Q-learning for different values of the cost $\eta$, \textit{before the agent has the key}. We can see that Q-learning in the original MDP ($\eta=0$) is attracted to the local optimum, while Q-learning in the lazy-MDP avoids the local optimum and eventually learns the optimal behavior (when $\eta=0.03$). \textbf{(d):} Scores (excluding the cost) of Q-learning with different values of $\eta$. Only the agent solving the lazy-MDP with $\eta=0.03$ converges to the optimal behavior. Both state occupancy measures and performance curves are averaged over 100 seeds.}
    \label{fig:kdt_exploration}
\end{figure}

A side effect of lazy-MDPs with a uniform random default policy is that they push agents to maintain randomness in states where acting randomly is affordable (\textit{i.e.} does not impact future performance too much). This is helpful in hard exploration tasks, where local minima make the exploration more difficult.
Actually, encouraging randomness in the policy via lazy-MDPs has two benefits: it regularizes behaviors and avoids determinism, and it rewards smart exploration where the random actions are only taken when it is safe to explore. To study the role of lazy-MDPs for exploration, we add a distractor state (\textit{i.e.} an absorbing state with a small reward) in the upper-left room in KDT so as to introduce a local minimum. Q-learning agents, even when explicitly increasing exploration (e.g. with linearly decayed epsilon-greedy action selection), mostly fail and always go for the distractor. In that situation, augmenting the MDP as a lazy-MDP with a uniform random default policy encourages random behaviors over consecutive steps, which helps exploring and going past the local minimum. We illustrate this effect on Fig.~\ref{fig:kdt_exploration}. For that experiment, we used tabular Q-learning with learning rate $\alpha=0.5$, epsilon-greedy exploration starting at $\epsilon_0=0.1$ and linearly decayed until $\epsilon_\infty=0$, $\gamma=0.99$, and a reward $r=0.1$ for the apple. Episodes that did not end in an absorbing state (\textit{i.e.} treasure or apple) were ended after 1000 steps. Under the right cost ($\eta=0.03$) lazy policies keep exploring after finding the small reward, and eventually find the key and the treasure, leading to a reward that justifies the cost for taking control. With a cost too high ($\eta=0.05$), lazy policies never take control.

\begin{figure}[t!]
    \centering
    \begin{subfigure}[b]{0.32\textwidth}
        \centering
        \includegraphics[width=\linewidth]{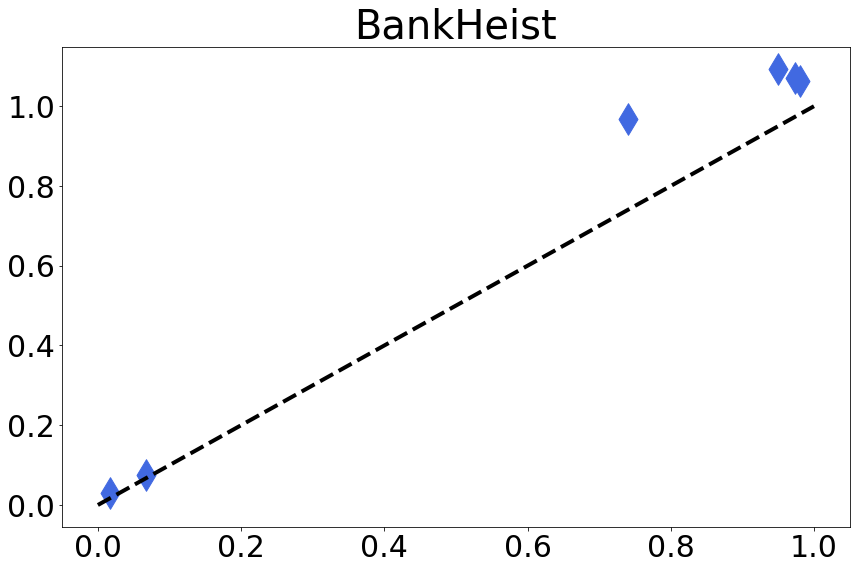}
      \end{subfigure}%
      \begin{subfigure}[b]{0.32\textwidth}
        \centering
        \includegraphics[width=\linewidth]{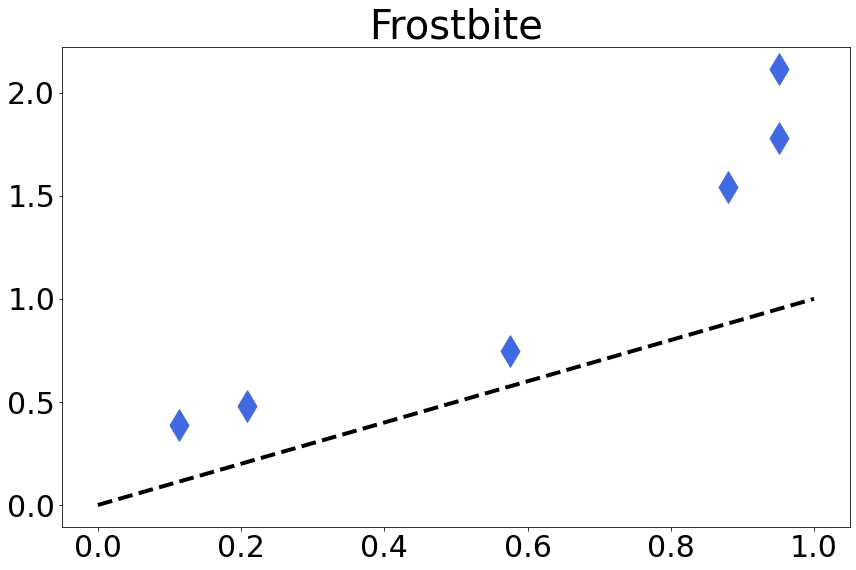}
      \end{subfigure}
     \begin{subfigure}[b]{0.32\textwidth}
        \centering
        \includegraphics[width=\linewidth]{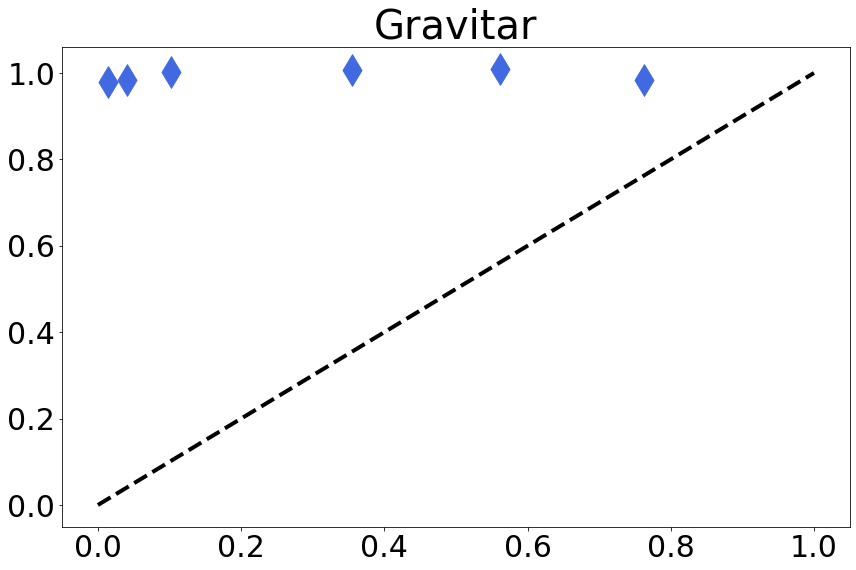}
      \end{subfigure}
      \begin{subfigure}[b]{0.32\textwidth}
        \centering
        \includegraphics[width=\linewidth]{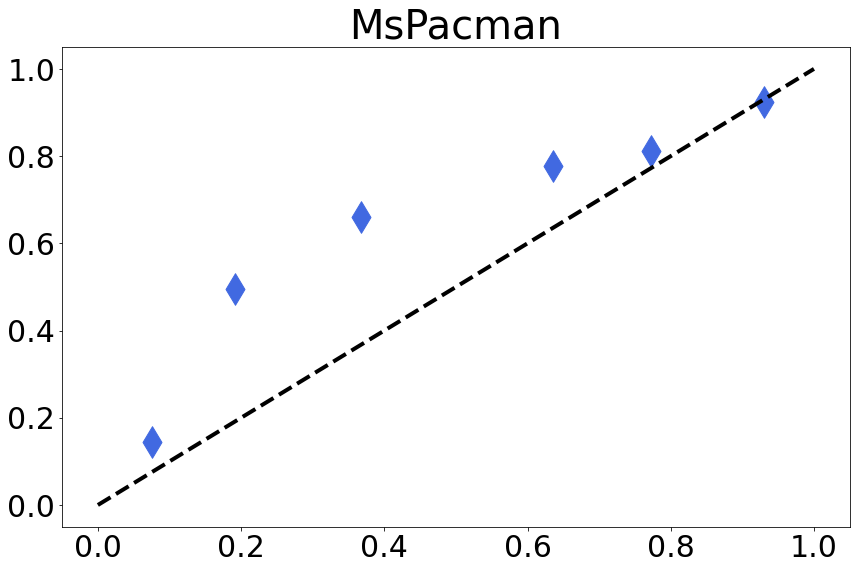}
      \end{subfigure}
      \begin{subfigure}[b]{0.32\textwidth}
        \centering
        \includegraphics[width=\linewidth]{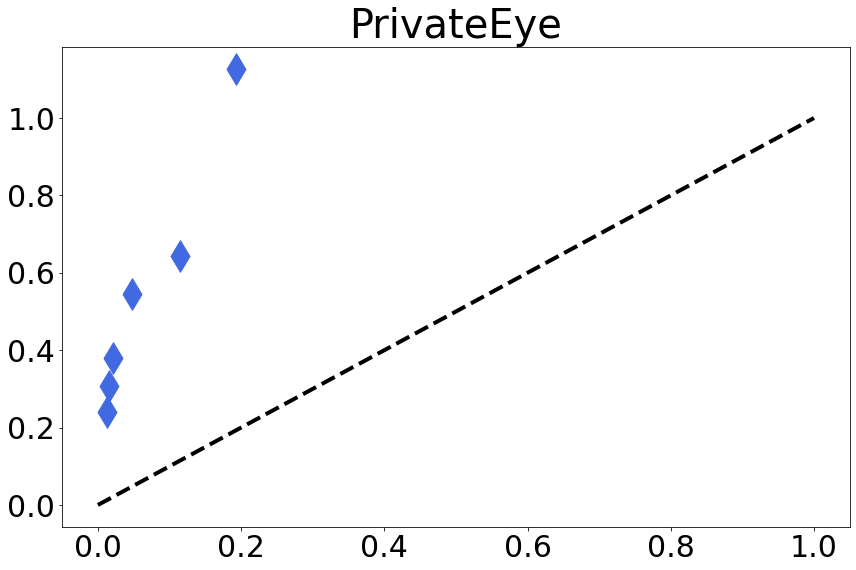}
      \end{subfigure}
      \begin{subfigure}[b]{0.32\textwidth}
        \centering
        \includegraphics[width=\linewidth]{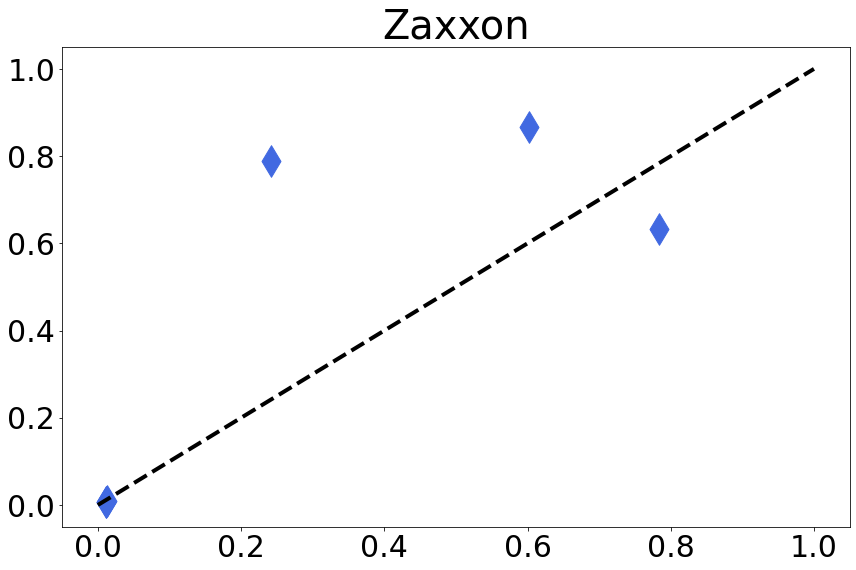}
      \end{subfigure}
    \caption{Percentage of the score of a standard DQN agent achieved by lazy-policies learned with different values of $\eta$ and uniformly random default policy on hard exploration tasks of Atari. 
    The x-axis represents the fraction of controls taken and the y-axis the percentage of the DQN baseline score reached.
    Each value is averaged over 3 seeds. Score 0\% correspond to getting no reward and score 100\% correspond to getting as much reward as a standard DQN.}
    \label{fig:atari_exploration}
\end{figure}

\looseness=-1
This motivates investigating if such an effect of taking less control while achieving higher returns can be observed in more complex games, requiring function approximation. Hence we converted hard exploration tasks in Atari to lazy-MDPs, including dense reward tasks (BankHeist, Frostbite, MsPacman, Zaxxon) and sparse reward tasks (Gravitar, PrivateEye), as classified in~\cite{bellemare2016unifying}. As previously, we used uniform random default policies and several values for the cost $\eta$ (0.005, 0.01, 0.02, 0.05, 0.1, 0.2) and reported the percentage of the score (with respect to a standard DQN agent~\citep{mnih2015humanlevel}) as a function of the resulting frequency of control taking (when the lazy policy does not choose the lazy action) in figure~\ref{fig:atari_exploration}. Reported values are averaged over 3 seeds. We observe that in most cases, reducing the frequency of control taking does not decrease the score too much (up to almost 100\% of the score in Gravitar with less than 10\% of control, 80\% of the score with less than 30\% of control in Zaxxon and more than 100\% of the score with 20\% of control on PrivateEye). Moreover, we even observed in Frostbite that the lazy policy learned with low penalties achieved 200\% of the score, confirming that lazy-MDP can also be used for improved exploration. 

\subsection{Learning on top of pretrained agents}
\label{subsec:residual}

\begin{figure*}[t]
    \centering
    \vspace{-2.5cm}
    \includegraphics[width=0.5\textwidth]{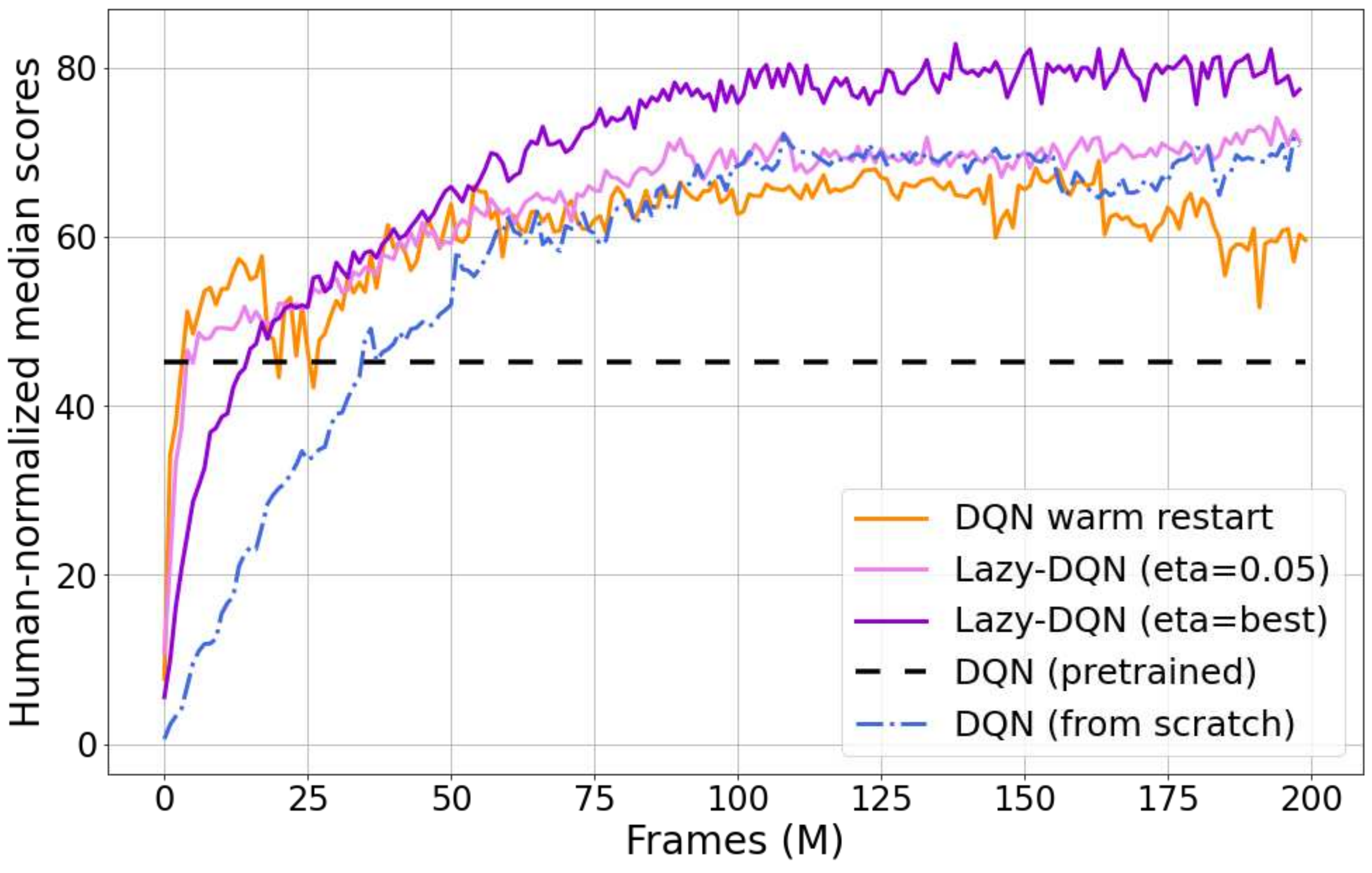}
    \caption{Human-normalized median scores in Atari (aggregated over the 60 games).}
    \label{fig:hnms}
\end{figure*}

The lazy-MDP framework can be viewed as an extension of residual learning to discrete action spaces.
Indeed, in lazy-MDPs, agents learn to sparsely take control over the default policy, in a subset of the states.
A natural question is: can they be used to reliably improve over the default policy?
To verify so, we measure the performance of DQN agents trained in the lazy-MDP version of Atari games, with a suboptimal DQN agent as default policy.
To ensure controlled suboptimality of the default policies, we use DQN agents at 50\% of their asymptotical performance.
Since learning in the lazy-MDPs benefits from the default policy (and the previous interaction it learned from), we compare it to warm-restarted DQN agents.
Warm restart consists in starting from pretrained weights instead of randomly initialized ones, and also benefits from past interactions.
Here, we use the weights of the agents taken as default policies in lazy-MDPs.

We measure the human-normalized median score~\citep{mnih2015humanlevel}, which aggregates performance across the 60 Atari 2600 games.
Results are reported in Fig.~\ref{fig:hnms}.
We notice that warm restarting is not very efficient: while performance increases faster than agents trained from scratch, it is also less stable and eventually decreases. 
Using lazy-MDPs with a good penalty ($\eta=0.05$), performance increases faster than DQN from scratch as well, with more stable improvements. There is no asymptotic gain on performance, which is not surprising given that we use the same agent. We observe the same phenomenon as in Sec.~\ref{subsec:exploration}: lazy-MDPs lead to better policies in hard exploration games (Frostbite, Zaxxon, Qbert, WizardOfWor, see Fig.~\ref{fig:all-atari}). 
A promising observation is that when selecting the best value of $\eta$ for each game, we observe a significant performance gain over from scratch DQN.
Also, lazy-MDPs can work with any type of policy as default policy (a program or a controller for instance), while warm restarting requires compatible weights.
We leave the study of automated $\eta$ selection to future work.
We display the performance curves in all 60 games in Fig.~\ref{fig:all-atari}.

\section{Conclusion}

In this work, we studied a novel paradigm for decision-making: learning when and how to act. We proposed lazy-MDPs, natural abstractions over MDPs that are well-suited for this learning problem, and showed that RL could still be used to provide solutions in that setting. We studied the theoretical properties of lazy-MDPs, including value functions and optimality. In experiments, we showed that lazy-MDPs present an interesting edge: when converted back to the original MDPs, the policies learned in lazy-MDPs tend to be more interpretable, as they highlight states where taking control is crucial to achieve increased returns.
With uniform random default policies, we show that policies learned in lazy-MDPs via DQN perform close to policies learned in standard MDPs, while only taking control in a fraction of the states; and can even reach higher scores in hard exploration tasks. With pretrained agents as default policies, policies learned in lazy-MDPs tend to outperform policies learned using warm restarts. 

\newpage

\balance
\bibliography{main}

\begin{thebibliography}{50}
\providecommand{\natexlab}[1]{#1}
\providecommand{\url}[1]{\texttt{#1}}
\expandafter\ifx\csname urlstyle\endcsname\relax
  \providecommand{\doi}[1]{doi: #1}\else
  \providecommand{\doi}{doi: \begingroup \urlstyle{rm}\Url}\fi

\bibitem[Amir and Amir(2018)]{amir2018highlights}
D.~Amir and O.~Amir.
\newblock Highlights: Summarizing agent behavior to people.
\newblock In \emph{International Conference on Autonomous Agents and Multiagent
  Systems}, 2018.

\bibitem[Amitai and Amir(2021)]{amitai2021disagreement}
Y.~Amitai and O.~Amir.
\newblock "{I} don't think so": Disagreement-based policy summaries for
  comparing agents.
\newblock \emph{arXiv preprint arXiv:2102.03064}, 2021.

\bibitem[Arjona-Medina et~al.(2019)Arjona-Medina, Gillhofer, Widrich,
  Unterthiner, Brandstetter, and Hochreiter]{arjona2019rudder}
J.~A. Arjona-Medina, M.~Gillhofer, M.~Widrich, T.~Unterthiner, J.~Brandstetter,
  and S.~Hochreiter.
\newblock Rudder: Return decomposition for delayed rewards.
\newblock In \emph{Advances in Neural Information Processing Systems}, 2019.

\bibitem[Bacon et~al.(2017)Bacon, Harb, and Precup]{bacon2017option}
P.-L. Bacon, J.~Harb, and D.~Precup.
\newblock The option-critic architecture.
\newblock In \emph{AAAI Conference on Artificial Intelligence}, 2017.

\bibitem[Barreto et~al.(2019)Barreto, Borsa, Hou, Comanici, Ayg{\"u}n, Hamel,
  Toyama, Hunt, Mourad, Silver, et~al.]{barreto2019option}
A.~Barreto, D.~Borsa, S.~Hou, G.~Comanici, E.~Ayg{\"u}n, P.~Hamel, D.~K.
  Toyama, J.~J. Hunt, S.~Mourad, D.~Silver, et~al.
\newblock The option keyboard: Combining skills in reinforcement learning.
\newblock In \emph{Advances in Neural Information Processing Systems}, 2019.

\bibitem[Bellemare et~al.(2016{\natexlab{a}})Bellemare, Srinivasan, Ostrovski,
  Schaul, Saxton, and Munos]{bellemare2016unifying}
M.~Bellemare, S.~Srinivasan, G.~Ostrovski, T.~Schaul, D.~Saxton, and R.~Munos.
\newblock Unifying count-based exploration and intrinsic motivation.
\newblock In \emph{Advances in neural information processing systems},
  2016{\natexlab{a}}.

\bibitem[Bellemare et~al.(2013)Bellemare, Naddaf, Veness, and
  Bowling]{bellemare2013arcade}
M.~G. Bellemare, Y.~Naddaf, J.~Veness, and M.~Bowling.
\newblock The arcade learning environment: An evaluation platform for general
  agents.
\newblock \emph{Journal of Artificial Intelligence Research}, 2013.

\bibitem[Bellemare et~al.(2016{\natexlab{b}})Bellemare, Ostrovski, Guez,
  Thomas, and Munos]{bellemare2016increasing}
M.~G. Bellemare, G.~Ostrovski, A.~Guez, P.~Thomas, and R.~Munos.
\newblock Increasing the action gap: New operators for reinforcement learning.
\newblock In \emph{Proceedings of the AAAI Conference on Artificial
  Intelligence}, 2016{\natexlab{b}}.

\bibitem[Bica et~al.(2021)Bica, Jarrett, H{\"u}y{\"u}k, and van~der
  Schaar]{bica2021learning}
I.~Bica, D.~Jarrett, A.~H{\"u}y{\"u}k, and M.~van~der Schaar.
\newblock Learning "what-if" explanations for sequential decision-making.
\newblock In \emph{International Conference on Learning Representations}, 2021.

\bibitem[Biedenkapp et~al.(2021)Biedenkapp, Rajan, Hutter, and
  Lindauer]{biedenkapp2020towards}
A.~Biedenkapp, R.~Rajan, F.~Hutter, and M.~Lindauer.
\newblock Towards {T}empo{RL}: learning when to act.
\newblock \emph{International Conference on Machine Learning, BIG workshop},
  2021.

\bibitem[Castro et~al.(2018)Castro, Moitra, Gelada, Kumar, and
  Bellemare]{castro2018dopamine}
P.~S. Castro, S.~Moitra, C.~Gelada, S.~Kumar, and M.~G. Bellemare.
\newblock Dopamine: {A} {R}esearch {F}ramework for {D}eep {R}einforcement
  {L}earning.
\newblock \emph{arXiv preprint arXiv:1812.06110}, 2018.
\newblock URL \url{http://arxiv.org/abs/1812.06110}.

\bibitem[Coppens et~al.(2019)Coppens, Efthymiadis, Lenaerts, Now{\'e}, Miller,
  Weber, and Magazzeni]{coppens2019distilling}
Y.~Coppens, K.~Efthymiadis, T.~Lenaerts, A.~Now{\'e}, T.~Miller, R.~Weber, and
  D.~Magazzeni.
\newblock Distilling deep reinforcement learning policies in soft decision
  trees.
\newblock In \emph{IJCAI/ECAI Workshop on Explainable Artificial Intelligence},
  2019.

\bibitem[Ferret et~al.(2019)Ferret, Marinier, Geist, and
  Pietquin]{ferret2019self}
J.~Ferret, R.~Marinier, M.~Geist, and O.~Pietquin.
\newblock Self-attentional credit assignment for transfer in reinforcement
  learning.
\newblock In \emph{International Joint Conference on Artificial Intelligence},
  2019.

\bibitem[Ferret et~al.(2021)Ferret, Pietquin, and Geist]{ferret2021self}
J.~Ferret, O.~Pietquin, and M.~Geist.
\newblock Self-imitation advantage learning.
\newblock In \emph{International Conference on Autonomous Agents and Multiagent
  Systems}, 2021.

\bibitem[Geist et~al.(2019)Geist, Scherrer, and Pietquin]{geist2019theory}
M.~Geist, B.~Scherrer, and O.~Pietquin.
\newblock A theory of regularized markov decision processes.
\newblock In \emph{International Conference on Machine Learning}, 2019.

\bibitem[Geist et~al.(2021)Geist, P{\'e}rolat, Lauri{\`e}re, Elie, Perrin,
  Bachem, Munos, and Pietquin]{geist2021concave}
M.~Geist, J.~P{\'e}rolat, M.~Lauri{\`e}re, R.~Elie, S.~Perrin, O.~Bachem,
  R.~Munos, and O.~Pietquin.
\newblock Concave utility reinforcement learning: the mean-field game
  viewpoint.
\newblock \emph{arXiv preprint arXiv:2106.03787}, 2021.

\bibitem[Greydanus et~al.(2018)Greydanus, Koul, Dodge, and
  Fern]{greydanus2018visualizing}
S.~Greydanus, A.~Koul, J.~Dodge, and A.~Fern.
\newblock Visualizing and understanding atari agents.
\newblock In \emph{International Conference on Machine Learning}, 2018.

\bibitem[Harutyunyan et~al.(2019)Harutyunyan, Dabney, Mesnard, Gheshlaghi~Azar,
  Piot, Heess, van Hasselt, Wayne, Singh, Precup,
  et~al.]{harutyunyan2019hindsight}
A.~Harutyunyan, W.~Dabney, T.~Mesnard, M.~Gheshlaghi~Azar, B.~Piot, N.~Heess,
  H.~P. van Hasselt, G.~Wayne, S.~Singh, D.~Precup, et~al.
\newblock Hindsight credit assignment.
\newblock In \emph{Advances in Neural Information Processing Systems}, 2019.

\bibitem[Huang et~al.(2019)Huang, Kavitha, and Zhu]{huang2019continuous}
Y.~Huang, V.~Kavitha, and Q.~Zhu.
\newblock Continuous-time markov decision processes with controlled
  observations.
\newblock In \emph{Allerton Conference on Communication, Control, and
  Computing}, 2019.

\bibitem[Hung et~al.(2019)Hung, Lillicrap, Abramson, Wu, Mirza, Carnevale,
  Ahuja, and Wayne]{hung2019optimizing}
C.-C. Hung, T.~Lillicrap, J.~Abramson, Y.~Wu, M.~Mirza, F.~Carnevale, A.~Ahuja,
  and G.~Wayne.
\newblock Optimizing agent behavior over long time scales by transporting
  value.
\newblock \emph{Nature communications}, 2019.

\bibitem[Johannink et~al.(2019)Johannink, Bahl, Nair, Luo, Kumar, Loskyll,
  Ojea, Solowjow, and Levine]{johannink2019residual}
T.~Johannink, S.~Bahl, A.~Nair, J.~Luo, A.~Kumar, M.~Loskyll, J.~A. Ojea,
  E.~Solowjow, and S.~Levine.
\newblock Residual reinforcement learning for robot control.
\newblock In \emph{International Conference on Robotics and Automation}, 2019.

\bibitem[Juozapaitis et~al.(2019)Juozapaitis, Koul, Fern, Erwig, and
  Doshi-Velez]{juozapaitis2019explainable}
Z.~Juozapaitis, A.~Koul, A.~Fern, M.~Erwig, and F.~Doshi-Velez.
\newblock Explainable reinforcement learning via reward decomposition.
\newblock In \emph{IJCAI/ECAI Workshop on Explainable Artificial Intelligence},
  2019.

\bibitem[Lakshminarayanan et~al.(2017)Lakshminarayanan, Sharma, and
  Ravindran]{lakshminarayanan2017dynamic}
A.~Lakshminarayanan, S.~Sharma, and B.~Ravindran.
\newblock Dynamic action repetition for deep reinforcement learning.
\newblock In \emph{AAAI Conference on Artificial Intelligence}, 2017.

\bibitem[Lee et~al.(2019)Lee, Eysenbach, Parisotto, Xing, Levine, and
  Salakhutdinov]{lee2019efficient}
L.~Lee, B.~Eysenbach, E.~Parisotto, E.~Xing, S.~Levine, and R.~Salakhutdinov.
\newblock Efficient exploration via state marginal matching.
\newblock \emph{arXiv preprint arXiv:1906.05274}, 2019.

\bibitem[Liu et~al.(2018)Liu, Schulte, Zhu, and Li]{liu2018toward}
G.~Liu, O.~Schulte, W.~Zhu, and Q.~Li.
\newblock Toward interpretable deep reinforcement learning with linear model
  u-trees.
\newblock In \emph{Joint European Conference on Machine Learning and Knowledge
  Discovery in Databases}, 2018.

\bibitem[Meresht et~al.(2020)Meresht, De, Singla, and
  Gomez-Rodriguez]{meresht2020learning}
V.~B. Meresht, A.~De, A.~Singla, and M.~Gomez-Rodriguez.
\newblock Learning to switch between machines and humans.
\newblock \emph{arXiv preprint arXiv:2002.04258}, 2020.

\bibitem[Mesnard et~al.(2021)Mesnard, Weber, Viola, Thakoor, Saade,
  Harutyunyan, Dabney, Stepleton, Heess, Guez,
  et~al.]{mesnard2021counterfactual}
T.~Mesnard, T.~Weber, F.~Viola, S.~Thakoor, A.~Saade, A.~Harutyunyan,
  W.~Dabney, T.~Stepleton, N.~Heess, A.~Guez, et~al.
\newblock Counterfactual credit assignment in model-free reinforcement
  learning.
\newblock In \emph{International Conference on Machine Learning}, 2021.

\bibitem[Mnih et~al.(2015)Mnih, Kavukcuoglu, Silver, Rusu, Veness, Bellemare,
  Graves, et~al.]{mnih2015humanlevel}
V.~Mnih, K.~Kavukcuoglu, D.~Silver, A.~A. Rusu, J.~Veness, M.~G. Bellemare,
  A.~Graves, et~al.
\newblock Human-level control through deep reinforcement learning.
\newblock \emph{Nature}, 2015.

\bibitem[Molnar(2020)]{molnar2020interpretable}
C.~Molnar.
\newblock \emph{Interpretable machine learning}.
\newblock Lulu. com, 2020.

\bibitem[Neu et~al.(2017)Neu, Jonsson, and G{\'o}mez]{neu2017unified}
G.~Neu, A.~Jonsson, and V.~G{\'o}mez.
\newblock A unified view of entropy-regularized markov decision processes.
\newblock \emph{arXiv preprint arXiv:1705.07798}, 2017.

\bibitem[Oh et~al.(2018)Oh, Guo, Singh, and Lee]{oh2018self}
J.~Oh, Y.~Guo, S.~Singh, and H.~Lee.
\newblock Self-imitation learning.
\newblock In \emph{International Conference on Machine Learning}, 2018.

\bibitem[Precup(2000)]{precup2000temporal}
D.~Precup.
\newblock \emph{Temporal abstraction in reinforcement learning}.
\newblock University of Massachusetts Amherst, 2000.

\bibitem[Puri et~al.(2019)Puri, Verma, Gupta, Kayastha, Deshmukh,
  Krishnamurthy, and Singh]{puri2019explain}
N.~Puri, S.~Verma, P.~Gupta, D.~Kayastha, S.~Deshmukh, B.~Krishnamurthy, and
  S.~Singh.
\newblock Explain your move: Understanding agent actions using specific and
  relevant feature attribution.
\newblock In \emph{International Conference on Learning Representations}, 2019.

\bibitem[Puterman(1994)]{puterman1994mdp}
M.~L. Puterman.
\newblock \emph{Markov Decision Processes}.
\newblock Wiley, 1994.

\bibitem[Raposo et~al.(2021)Raposo, Ritter, Santoro, Wayne, Weber, Botvinick,
  van Hasselt, and Song]{raposo2021synthetic}
D.~Raposo, S.~Ritter, A.~Santoro, G.~Wayne, T.~Weber, M.~Botvinick, H.~van
  Hasselt, and F.~Song.
\newblock Synthetic returns for long-term credit assignment.
\newblock \emph{arXiv preprint arXiv:2102.12425}, 2021.

\bibitem[Rummery and Niranjan(1994)]{rummery1994online}
G.~A. Rummery and M.~Niranjan.
\newblock \emph{Online Q-learning using connectionist systems}, volume~37.
\newblock Citeseer, 1994.

\bibitem[Shani et~al.(2019)Shani, Efroni, and Mannor]{shani2019exploration}
L.~Shani, Y.~Efroni, and S.~Mannor.
\newblock Exploration conscious reinforcement learning revisited.
\newblock In \emph{International Conference on Machine Learning}, 2019.

\bibitem[Sharma et~al.(2017)Sharma, Lakshminarayanan, and
  Ravindran]{sharma2017learning}
S.~Sharma, A.~S. Lakshminarayanan, and B.~Ravindran.
\newblock Learning to repeat: Fine grained action repetition for deep
  reinforcement learning.
\newblock In \emph{International Conference on Learning Representations}, 2017.

\bibitem[Silver et~al.(2018)Silver, Allen, Tenenbaum, and
  Kaelbling]{silver2018residual}
T.~Silver, K.~Allen, J.~Tenenbaum, and L.~Kaelbling.
\newblock Residual policy learning.
\newblock \emph{arXiv preprint arXiv:1812.06298}, 2018.

\bibitem[Sutton and Barto(2018)]{sutton1998reinforcement}
R.~S. Sutton and A.~G. Barto.
\newblock \emph{Reinforcement Learning: An Introduction}.
\newblock The MIT Press, 2018.

\bibitem[Sutton et~al.(1999)Sutton, Precup, and Singh]{sutton1999between}
R.~S. Sutton, D.~Precup, and S.~Singh.
\newblock Between mdps and semi-mdps: A framework for temporal abstraction in
  reinforcement learning.
\newblock \emph{Artificial intelligence}, 1999.

\bibitem[Topin and Veloso(2019)]{topin2019generation}
N.~Topin and M.~Veloso.
\newblock Generation of policy-level explanations for reinforcement learning.
\newblock In \emph{AAAI Conference on Artificial Intelligence}, 2019.

\bibitem[Torrey and Taylor(2013)]{torrey2013teaching}
L.~Torrey and M.~Taylor.
\newblock Teaching on a budget: Agents advising agents in reinforcement
  learning.
\newblock In \emph{Proceedings of the 2013 international conference on
  Autonomous agents and multi-agent systems}, 2013.

\bibitem[Van~Seijen et~al.(2009)Van~Seijen, Van~Hasselt, Whiteson, and
  Wiering]{van2009theoretical}
H.~Van~Seijen, H.~Van~Hasselt, S.~Whiteson, and M.~Wiering.
\newblock A theoretical and empirical analysis of expected sarsa.
\newblock In \emph{IEEE symposium on adaptive dynamic programming and
  reinforcement learning}, pages 177--184, 2009.

\bibitem[Verma et~al.(2019{\natexlab{a}})Verma, Le, Yue, and
  Chaudhuri]{verma2019imitation}
A.~Verma, H.~M. Le, Y.~Yue, and S.~Chaudhuri.
\newblock Imitation-projected programmatic reinforcement learning.
\newblock In \emph{Advances in Neural Information Processing Systems},
  2019{\natexlab{a}}.

\bibitem[Verma et~al.(2019{\natexlab{b}})Verma, Murali, Singh, Kohli, and
  Chaudhuri]{verma2018programatically}
A.~Verma, V.~Murali, R.~Singh, P.~Kohli, and S.~Chaudhuri.
\newblock Programmatically interpretable reinforcement learning.
\newblock In \emph{International Conference on Machine Learning},
  2019{\natexlab{b}}.

\bibitem[Vieillard et~al.(2020)Vieillard, Kozuno, Scherrer, Pietquin, Munos,
  and Geist]{vieillard2020leverage}
N.~Vieillard, T.~Kozuno, B.~Scherrer, O.~Pietquin, R.~Munos, and M.~Geist.
\newblock Leverage the average: an analysis of kl regularization in rl.
\newblock In \emph{Advances in Neural Information Processing Systems}, 2020.

\bibitem[Wang et~al.(2015)Wang, Schaul, Hessel, Van~Hasselt, Lanctot, and
  De~Freitas]{wang2015dueling}
Z.~Wang, T.~Schaul, M.~Hessel, H.~Van~Hasselt, M.~Lanctot, and N.~De~Freitas.
\newblock Dueling network architectures for deep reinforcement learning.
\newblock In \emph{International Conference on Machine Learning}, 2015.

\bibitem[Watkins and Dayan(1992)]{watkins1992q}
C.~J. Watkins and P.~Dayan.
\newblock Q-learning.
\newblock \emph{Machine learning}, 1992.

\bibitem[Zahavy et~al.(2016)Zahavy, Ben-Zrihem, and Mannor]{zahavy2016graying}
T.~Zahavy, N.~Ben-Zrihem, and S.~Mannor.
\newblock Graying the black box: Understanding dqns.
\newblock In \emph{International Conference on Machine Learning}, 2016.

\end{thebibliography}
\bibliographystyle{abbrvnat}

\newpage
\appendix
\onecolumn

\section{Proof of Thm.~\ref{th1}}\label{appendix:th1}

\begin{theorem*}%
    The cost function satisfies the following Bellman equation:
    \begin{equation}
        C^{\pi_+}(s) = -\eta(1- \pi_+(\bar{a}\vert s)) + \gamma \mathbb{E}_{a\sim \pi, s'\sim \mathcal{P}(\cdot \vert s,a)} C^{\pi_+}(s'). 
    \end{equation}
\end{theorem*}

\begin{proof}
\begin{align}
    V_+^{\pi_+}(s) &\eqdef \sum\limits_{a\in\mathcal{A}_+}\pi_+(a\vert s)\bigg(r_+(s,a) + \gamma \mathbb{E}_{s'}\Big[V_+^{\pi_+}(s')\Big]\bigg)\\
    &= \sum\limits_{a\in \mathbb{A}} \Big( \pi(a\vert s) - \pi_+(\bar{a}\vert s)\bar{\pi}(a\vert s)\Big) \bigg( r(s,a) - \eta +\gamma \mathbb{E}_{s'}\Big[V_+^{\pi_+}(s)\Big]\bigg) \\
    &\hspace{5mm}+ \pi_+(\bar{a}\vert s) \mathbb{E}_{\bar{\pi}}\bigg[ r(s,a) +\gamma \mathbb{E}_{s'}\Big[V_+^{\pi_+}(s)\Big]\bigg]\\
    &= \sum\limits_{a\in \mathbb{A}} \pi(a\vert s) \bigg( r(s,a) - \eta +\gamma \mathbb{E}_{s'}\Big[V_+^{\pi_+}(s)\Big]\bigg)\\
    &\hspace{5mm}-
    \pi_+(\bar{a}\vert s) \sum\limits_{a\in \mathbb{A}}\bar{\pi}(a\vert s) \bigg( r(s,a) - \eta +\gamma \mathbb{E}_{s'}\Big[V_+^{\pi_+}(s)\Big]\bigg)\\
    &\hspace{5mm} + 
    \pi_+(\bar{a}\vert s) \mathbb{E}_{\bar{\pi}}\bigg[ r(s,a) +\gamma \mathbb{E}_{s'}\Big[V_+^{\pi_+}(s)\Big]\bigg]\\
    &= \sum\limits_{a\in \mathbb{A}} \pi(a\vert s) \bigg( r(s,a) +\gamma \mathbb{E}_{s'}\Big[V_+^{\pi_+}(s)\Big]\bigg) - \eta + \pi_+(\bar{a}\vert s) \eta.
\end{align}
So,
\begin{align}
    V^\pi(s) + C^{\pi_+}(s) &= \mathbb{E}_{a\sim \pi, s'\sim s,a } \bigg( r(s,a) +\gamma \mathbb{E}_{s'}\Big[V^\pi(s') + C^{\pi_+}(s')\Big]\bigg)\\
    &- \eta( 1-\pi_+(\bar{a}\vert s)),
\end{align}
$\Leftrightarrow$
\begin{equation}
    C^{\pi_+}(s) = - \eta( 1-\pi_+(\bar{a}\vert s)) +\gamma \mathbb{E}_{a\sim \pi, s'\sim s,a }\Big[C^{\pi_+}(s')\Big],
\end{equation}
as
\begin{equation}
    V^\pi(s) = \mathbb{E}_{a\sim \pi, s'\sim s,a } \bigg( r(s,a) +\gamma \mathbb{E}_{s'}\Big[V^\pi(s')\Big]\bigg).
\end{equation}
\end{proof}

\section{Proof of Prop.~\ref{p1}}

\begin{prop}
\begin{align}
    V_+^{\pi_+}(s) =& (1 - \pi_+(\bar{a}\vert s)) \mathbb{E}_{a\sim\pi_{\setminus \bar{a}}}\Big[Q_{\setminus \bar{a}}^{\pi_+}(s,a)\Big]\\ 
    &+ \pi_+(\bar{a}\vert s) \bigg(\mathbb{E}_{a\sim\bar{\pi}}\Big[Q_{\setminus \bar{a}}^{\pi_+}(s,a)\Big] + \eta\bigg).
\end{align}
\end{prop}

\begin{proof}
We separately compute $V^\pi$ and $C^{\pi_+}$, the two components of $V_+^{\pi_+}$. First we have:
\begin{align}
    V^\pi(s) &= \sum\limits_{a\in\mathcal{A}}\pi(a\vert s)Q^\pi(s,a)\\
    &= \sum\limits_{a\in\mathcal{A}}\pi_+(a\vert s)Q^\pi (s,a) + \pi_+(\bar{a}\vert s)\sum\limits_{a\in\mathcal{A}}\bar{\pi}(a\vert s)Q^\pi(s,a)\\
    &= (1-\pi_+(\bar{a}\vert s))\mathbb{E}_{a\sim \pi_{\setminus \bar{a}}}\bigg[Q^\pi(s,a)\bigg] + \pi_+(\bar{a}\vert s)\mathbb{E}_{a\sim \bar{\pi}}\Big[Q^\pi(s,a)\Big].
\end{align}
Second, we have:
\begin{align}
    C^{\pi_+}(s) &= -(1-\pi_+(\bar{a}\vert s))\eta + \gamma\sum\limits_{a\in\mathcal{A}}\pi(a\vert s)\mathbb{E}_{s'}\Big[C^{\pi_+}(s')\Big]\\
    &= -(1-\pi_+(\bar{a}\vert s))\eta \\
    &\hspace{5mm}+\gamma\sum\limits_{a\in\mathcal{A}} \Big(\pi_+(a\vert s) + \pi_+(\bar{a} \vert s) \bar{\pi}(a\vert s)\Big)\mathbb{E}_{s'}\Big[C^{\pi_+}(s')\Big]\\
    &= (1-\pi_+(\bar{a}\vert s))\bigg(-\eta+\gamma\mathbb{E}_{a\sim \pi_{\setminus \bar{a}}}\Big[C^{\pi_+}(s')\Big]\bigg) \\
    &\hspace{5mm}+ \pi_+(\bar{a}\vert s)\gamma\mathbb{E}_{a\sim \bar{\pi}}\Big[C^{\pi_+}(s')\Big].
\end{align}
Consequently,
\begin{align}
    V_+^{\pi_+}(s) &= V^\pi(s) + C^{\pi_+}(s) \textit{ (from Theorem~\ref{th1})}\\
    &= (1 - \pi_+(\bar{a}\vert s)) \mathbb{E}_{a\sim\pi_{\setminus \bar{a}}}\Big[Q_{\setminus \bar{a}}^{\pi_+}(s,a)\Big] \\
    &\hspace{5mm}+ \pi_+(\bar{a}\vert s) \bigg(\mathbb{E}_{a\sim\bar{\pi}}\Big[Q_{\setminus \bar{a}}^{\pi_+}(s,a)\Big] + \eta\bigg).
\end{align}
\end{proof}

\section{Proof of Prop.~\ref{p2}}

\begin{prop}
\begin{align}
    Q^{\pi_+}_+(s,a) &=
    \begin{cases}
        Q_{\setminus \bar{a}}^{\pi_+}(s,a) &\text{if } a\neq\bar{a},\\
        \mathbb{E}_{a\sim\bar{\pi}}\Big[Q_{\setminus \bar{a}}^{\pi_+}(s,a)\Big] +\eta &\text{if } a=\bar{a}.
    \end{cases}
\end{align}
\end{prop}
\begin{proof}
\begin{align}
    Q^{\pi_+}_+(s,a) &\eqdef r_+(s,a) +\gamma\mathbb{E}_{s'\sim \mathcal{P}_+(.\vert s,a)}\Big[V^{\pi_+}_+(s')\Big]\\
    &=
    \begin{cases}
        r(s,a) - \eta + \gamma\mathbb{E}_{s'\sim \mathcal{P}(.\vert s,a)}\Big[V^{\pi_+}_+(s')\Big] &\text{if } a\neq\bar{a}\\
        \mathbb{E}_{a\sim\bar{\pi}}\bigg[r(s,a) + \gamma\mathbb{E}_{s'\sim \mathcal{P}(.\vert s,a)}\Big[V^{\pi_+}_+(s')\Big]\bigg] &\text{if } a=\bar{a}
    \end{cases}\\
    &=
    \begin{cases}
        Q_{\setminus \bar{a}}^{\pi_+}(s,a) &\text{if } a\neq\bar{a}\\
        \mathbb{E}_{a\sim\bar{\pi}}\Big[Q_{\setminus \bar{a}}^{\pi_+}(s,a)\Big] +\eta &\text{if } a=\bar{a}
    \end{cases}.
\end{align}
\end{proof}

\section{Proof of Prop.~\ref{p3}}

\begin{prop}
The following policy $\pi_+$ is greedy with respect to $Q_+$:
\begin{align}
    \pi_+(\cdot \vert s) &= 
    \begin{cases}
    \frac{\mathbb{I}\Big\{a\in\argmax_\mathcal{A} Q_{+\setminus\bar{a}}(s,a)\Big\}}{\Big\lvert \argmax_\mathcal{A} Q_{+\setminus\bar{a}}(s,a) \Big\rvert} &\text{if } G_{Q_{+\setminus\bar{a}}}(s)>\eta,\\
    \mathbb{I}\Big\{a=\bar{a}\Big\} &\text{otherwise}.
    \end{cases}
\end{align}
\end{prop}
\begin{proof}
\begin{align}
    \max_\mathcal{A}Q_+(s,\cdot) > Q_+(s,\bar{a}) &\Leftrightarrow \max_\mathcal{A}Q_{+\setminus\bar{a}}(s,a) > \mathbb{E}_{\bar{\pi}}\Big[Q_{+\setminus\bar{a}}(s,a)\Big]+\eta\\ 
    &\Leftrightarrow G_{Q_{+\setminus\bar{a}}}(s)>\eta.
\end{align}
\end{proof}

\section{Proof of Thm.~\ref{th:convergence}}\label{appendix:convergence}

\begin{theorem*}
    $\mathcal{T}$ is a $\gamma$-contraction, and converges to $Q_{\setminus \bar{a}}^{\pi^*_+}$ where $\pi^*_+$ is the optimal policy in the lazy-MDP.
\end{theorem*}
\begin{proof}
First, we define the operator in the augmented MDP
\begin{equation}
    \mathcal{T_+}:Q_+(s,a) \rightarrow r_+(s,a) + \gamma\mathbb{E}_{s'\sim \mathcal{P}_+(.\vert s,a)}\Big[\max_{\mathcal{A}_+}Q_+(s', .)\Big]
\end{equation}
which is known to be a contraction, and its successive applications converges to the unique, optimal fixed point, $Q_+^*(s,a\in \mathcal{A}_+)$:
\begin{equation}
    Q_+^*(s,a) = r_+(s,a) + \gamma\mathbb{E}_{s'\sim \mathcal{P}_+(.\vert s,a)}\Big[\max_{\mathcal{A}_+}Q^*_+(s', .)\Big].
\end{equation}

    Now, let $Q_{{\setminus \bar{a}}_1}(s, a\in \mathcal{A})$ and $Q_{{\setminus \bar{a}}_2}(s, a\in \mathcal{A})$ be arbitrary Q-function of the base MDP, let $s,a$ be any state-action pair in the base MDP, and assume without loss of generality that $T Q_{{\setminus \bar{a}}_1}(s,a) \geq T Q_{{\setminus \bar{a}}_2}(s,a)$. We have
    \begin{align}
        0 &\leq |\mathcal{T}Q_{{\setminus \bar{a}}_1}(s,a) - \mathcal{T}Q_{{\setminus \bar{a}}_2}(s,a)|
        \\
        &= \mathcal{T}Q_{{\setminus \bar{a}}_1}(s,a) - \mathcal{T}Q_{{\setminus \bar{a}}_2}(s,a)
        \\
        &= \mathcal{T}_+ Q_{1+}(s,a) - \mathcal{T}_+ Q_{2+}(s,a)
        \\
        &= \gamma \mathbb{E}_{s'|s,a}[Q_{1+}(s',a_1^*) - \underbrace{Q_{2+}(s',a_2^*)}_{\geq Q_{2+}(s',a_1^*)}]
        \text{ with } a_i^* \in \argmax_{\mathcal{A}^+} Q_i(s',\cdot)
        \\
        &\leq \gamma \mathbb{E}_{s'|s,a}[Q_{1+}(s',a_1^*) - Q_{2+}(s',a_1^*)].
    \end{align}
    If $a_1^* \neq \bar{a}$, we have
    \begin{align}
        \gamma \mathbb{E}_{s'|s,a} \Big[Q_{1+}(s',a_1^*) &- Q_{2+}(s',a_1^*) \Big]\\
        &=  \gamma \mathbb{E}_{s'|s,a}\Big[Q_{{\setminus \bar{a}}_1}(s',a_1^*) - Q_{{\setminus \bar{a}}_{2+}}(s',a_1^*) \Big]
        \\
        &\leq \Big\|Q_{{\setminus \bar{a}}_1} - Q_{{\setminus \bar{a}}_2} \Big\|_\infty.
    \end{align}
    If $a_1^* = \bar{a}$, we have
    \begin{align}
        &\gamma \mathbb{E}_{s'|s,a} \Big[Q_{1+}(s',a_1^*) - Q_{2+}(s',a_1^*) \Big]\\
        &\hspace{5mm}=  \gamma \mathbb{E}_{s'|s,a}\Big[\mathbb{E}_{\bar{\pi}} \big[\hat{Q_1}(s',a') + \eta \big]-\mathbb{E}_{\bar{\pi}}\big[\hat{Q_2}(s',a') + \eta \big]\Big]
        \\
        &\hspace{5mm}\leq \Big\|Q_{{\setminus \bar{a}}_1} - Q_{{\setminus \bar{a}}_2} \Big\|_\infty.
    \end{align}
    Thus, we can conclude that $\|\mathcal{T}Q_{{\setminus \bar{a}}_1} - \mathcal{T}Q_{{\setminus \bar{a}}_2}\|_\infty\leq \gamma \|Q_{{\setminus \bar{a}}_1} - Q_{{\setminus \bar{a}}_2}\|_\infty$, making $\mathcal{T}$ a $\gamma$-contraction. \\
\\
Now we show that successively applied, it converges to $Q_{\setminus \bar{a}}^{\pi^*_+}$.\\
\\
Let $\mathcal{T}_+^N(q_+) = \underbrace{\mathcal{T}_+ \circ \hdots \circ \mathcal{T}_+}_{N \text{times}}(Q_+)$. As $\mathcal{T}_+^N(Q_+)$ converges to $Q_+^*$ with $N$, we have:
\begin{equation}
    \forall \epsilon>0,\hspace{3mm}\exists N_\epsilon,\hspace{3mm}
    \norm{\mathcal{T}_+^{N_\epsilon}(Q_+) - Q_+^*}_\infty < \epsilon
\end{equation}
Besides, for $a\neq\bar{a}$, we have:
\begin{align}
    \mathcal{T_+}(Q_+)(s,a) &= \mathcal{T}(Q_{\setminus \bar{a}})(s,a),\\
    Q_+^*(s,a) &= Q_{\setminus \bar{a}}^*(s,a),
\end{align}
so,
\begin{align}
    \forall s\in\mathcal{S}, a\in\mathcal{A}\hspace{3mm}
    \Big\vert \mathcal{T}^{N_\epsilon}(Q_{\setminus \bar{a}})(s,a) &- Q_{\setminus \bar{a}}^*(s,a) \Big\vert =\\
    &\Big\vert \mathcal{T}_+^{N_\epsilon}(Q_+)(s,a) - Q_+^*(s,a) \Big\vert,
\end{align}
and
\begin{align}
    \norm{\mathcal{T}^{N_\epsilon}(Q_{\setminus \bar{a}}) - Q_{\setminus \bar{a}}^*}_\infty &=
    \max\limits_{s\in\mathcal{S}, a\in\mathcal{A}}
    \Big\vert \mathcal{T}^{N_\epsilon}(Q_{\setminus \bar{a}})(s,a) - Q_{\setminus \bar{a}}^*(s,a) \Big\vert\\
    &=
    \max\limits_{s\in\mathcal{S}, a\in\mathcal{A}} \Big \vert \mathcal{T}_+^{N_\epsilon}(Q_+)(s,a) - Q_+^*(s,a)\Big\vert\\
    &\leq \max\limits_{s\in\mathcal{S}, a\in\mathcal{A}_+} \Big \vert \mathcal{T}_+^{N_\epsilon}(Q_+)(s,a) - Q_+^*(s,a)\Big\vert\\
    &= \norm{\mathcal{T}_+^{N_\epsilon}(Q_+) - Q_+^*}_\infty < \epsilon
\end{align}
\end{proof}

\section{Proof of Thm.~\ref{th3}}

\begin{theorem}
Let $Q^{\bar{\pi}}(s, a\in \mathcal{A})$ be the Q-function of the default policy in the base MDP. Then: $\eta_\text{max} = \max_s G_{Q^{\bar{\pi}}}(s)$.
\end{theorem}
\begin{proof}
Since the lazy-gap does not depend on $\eta$ when the agent always follows the default policy:
\begin{equation}
    \eta_{\max} = \sup\Big\{\eta>0 \, \vert\hspace{2mm} \exists s,\hspace{2mm} G^*(s)=\eta \Big\} = \max\limits_s G^*(s).
\end{equation}
\end{proof}

\section{Proof of Thm.\ref{th:eta_min}}
\label{appendix:proofs}

\begin{theorem*}
Let $\pi^*$ be the optimal policy in the base MDP, and $Q^*(s, a\in \mathcal{A})$ the associated Q-function. Then:
\begin{align}
    &\eta_\text{min} =\\ 
    &\hspace{2mm} \min_s\max_a \frac{Q^*(s,a) - \mathbb{E}_{\bar{\pi}}\Big[Q^*(s, \cdot)\Big]}{ 1 + \bigg(\mathbb{E}_{\pi^*}\Big[Z^{\pi^*}(s, \cdot)\Big] - \mathbb{E}_{\bar{\pi}}\Big[Z^{\pi^*}(s, \cdot)\Big]\bigg)},
\end{align}
with $\eta_\text{min} \geq 0$.
\end{theorem*}

\begin{proof}
\begin{cor}
Let $x(t) = \frac{u(t)}{1+v(t)}$ for all $t$ where $x(t)\geq0$, and $x_\text{min} = \min_t x(t)$, $x_\text{max} = \max_t x(t)$ Then:
\begin{align}
    x_\text{min} = \min_t\Big( u(t) - x_\text{min} v(t)\Big)\\
    x_\text{max} = \max_t\Big( u(t) - x_\text{max} v(t)\Big)
\end{align}
\end{cor}

Proof of corollary:\\
We have, for all $t$, $x_\text{min}=x(t^*)\leq x(t)$.
Since $x(t)>0$, we have for all $t$,
\begin{align}
    u(t) - x(t^*)v(t)\geq u(t) - x(t)v(t) = x(t)
\end{align}
And since $x(t)>x(t^*)$, for all $t$, 
\begin{equation}
    u(t) - x(t^*)v(t)\geq x(t^*) = u(t^*) - x(t^*)v(t^*),
\end{equation}
which means that 
\begin{equation}
    x(t^*)=\min_t \Big(u(t) - x(t^*)v(t)\Big)
\end{equation}

Similarly, we have, for all $t$, $x_\text{max}=x(T^*)\geq x(t)$.
Since $x(t)>0$, we have for all $t$,
\begin{align}
    u(t) - x(T^*)v(t)\leq u(t) - x(t)v(t) = x(t)
\end{align}
And since $x(t)\leq x(T^*)$, for all $t$, 
\begin{equation}
    u(t) - x(T^*)v(t)\leq x(T^*) = u(T^*) - x(T^*)v(T^*),
\end{equation}
which means that 
\begin{equation}
    x(T^*)=\max_t \Big(u(t) - x(T^*)v(t)\Big)
\end{equation}
So we prove both qualities of the corollary.
An immediate consequence of this corollary is that if $x(s,t) = \frac{u(s,t)}{1+v(s,t)}$ with $v(s,t)\geq0$, then $x_\text{minimax} = \min_s\max_t x(s,t)$ verifies:
\begin{equation}\label{minimax}
    x_\text{minimax} = \min_s\max_t \Big(u(s,t) - x_\text{minimax}v(s,t)\Big)
\end{equation}

Now, we will use $u(s,a) = Q^*(s,a) - \mathbb{E}_{\bar{\pi}}\Big[Q^*(s,.)\Big]$ and $v(s,a) = \mathbb{E}_{\pi^*}\Big[Z^{\pi^*}(s,.)\Big] - \mathbb{E}_{\bar{\pi}}\Big[Z^{\pi^*}(s,.)\Big]$. Note that $u(s,a)>0$. If $v(s,a)\leq -1$ then whatever is $\eta$, it is always better to follow $\pi^*$ (indeed, the lazy-gap is always larger than $\eta$):
\begin{align}
    G_{Q_{\setminus \bar{a}}^*}(s) &= \max_a Q_{\setminus \bar{a}}^*(s,a) - \mathbb{E}_{\bar{\pi}}\Big[Q_{\setminus \bar{a}}^*(s,.)\Big]\\
    &= \max_a Q^{\pi^*}(s,a) - \eta \mathbb{E}_{\pi^*}\Big[Z^{\pi^*}(s,.)\Big]
    - \mathbb{E}_{\bar{\pi}}\Big[Q^{\pi^*}(s,.) - \eta Z^{\pi^*}(s,.)\Big]\\
    &= \max_a u(s,a) - \eta v(s,a)
    > \eta
\end{align}
Hence we only consider $v(s,a) > -1$. In that case, $x(s,a) = \frac{u(s,a)}{1+v(s,a)}>0$ and we can apply Eq~\eqref{minimax} to $\eta^*= x_\text{minimax}$:
\begin{align}
    \eta^* &= \min_s\max_a \Big(u(s,a) - \eta^*v(s,a)\Big)\\
    &= \min_s G_{Q_{\setminus \bar{a}}^*}(s)
\end{align}

\end{proof}

\section{Effect of the cost}\label{appendix:eta_min_max}
\begin{figure}[t]
    \centering
    \begin{subfigure}[b]{0.24\linewidth}
        \centering
        \includegraphics[width=\linewidth]{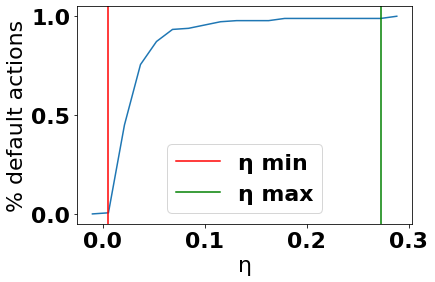}
        \caption{}
    \end{subfigure}
    \begin{subfigure}[b]{0.24\linewidth}
        \centering
        \includegraphics[width=\linewidth]{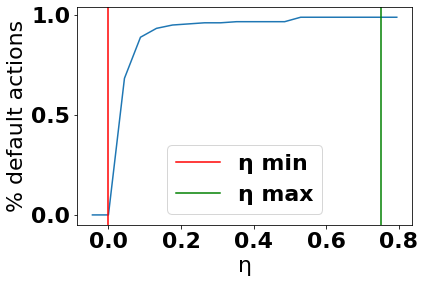}
        \caption{}
    \end{subfigure}
    \begin{subfigure}[b]{0.24\linewidth}
        \centering
        \includegraphics[width=\linewidth]{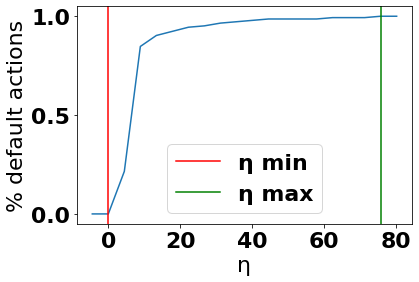}
        \caption{}
    \end{subfigure}
    \begin{subfigure}[b]{0.24\linewidth}
        \centering
        \includegraphics[width=\linewidth]{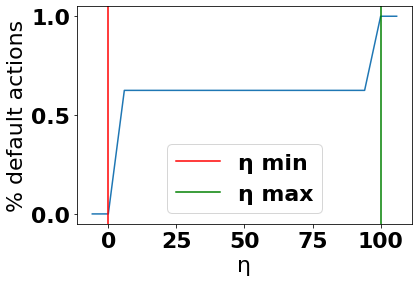}
        \caption{}
    \end{subfigure}
    \caption{\looseness=-1 Frequency of the lazy action as a function of $\eta$ in various scenarios. With $\eta < \eta_{\min}$, the agent never uses the lazy action, and with $\eta > \eta_{\max}$, the agent always uses the lazy action. Note that the empirical values for $\eta_{\min}$ and $\eta_{\max}$ match the values calculated with the theoretical expressions from Sec.~\ref{eta_min_max}. \textbf{(a)} KDT with a uniform random default policy. \textbf{(b)} KDT with a default policy that always takes second-best actions. \textbf{(c)} R\&B with a uniform random default policy. \textbf{(d)} R\&B with a default policy that always takes second-best actions.}
    \label{fig:eta}
\end{figure}

In order to empirically validate the cost boundaries we established in Sec.~\ref{eta_min_max}, we display in Fig.~\ref{fig:eta} the frequency of lazy actions played by an optimal agent as a function of $\eta$ in the R\&B and KDT environments, under different default policies. As expected, when $\eta < \eta_{\min}$ no lazy actions are ever selected, and when $\eta > \eta_{\max}$, the agent always chooses the lazy action. 

\section{Implementation details}\label{implementation_details}

For the Atari experiments, we used the default architectures and hyperparameters in the standard implementations from Dopamine. Details of the networks architecture, learning procedure and hyperparameter selection are described in~\citet{castro2018dopamine}.
For the discrete environments (KDT and R\&B) we could compute exact value functions via Value Iteration up to convergence. In the exploration experiment, we assumed that the agent had no knowledge of the dynamics of the environment and used tabular Q-learning with learning rate $\alpha=0.5$, epsilon-greedy exploration starting at $\epsilon_0=0.1$ and linearly decayed until $\epsilon_\infty=0$. The training is performed over 100 steps containing 1000 episodes of maximum length 1000. %

\section{Lazy-gap as a measure of state importance}\label{annex:importance}

\begin{figure*}[ht!]
    \centering
    \begin{subfigure}[b]{0.24\linewidth}
        \centering
        \includegraphics[width=\linewidth]{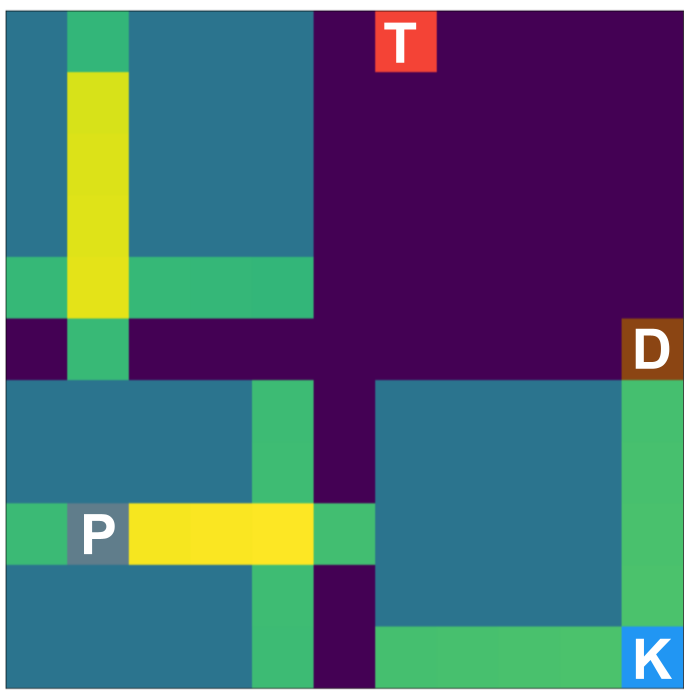}
      \end{subfigure}%
      \begin{subfigure}[b]{0.24\linewidth}
        \centering
        \includegraphics[width=\linewidth]{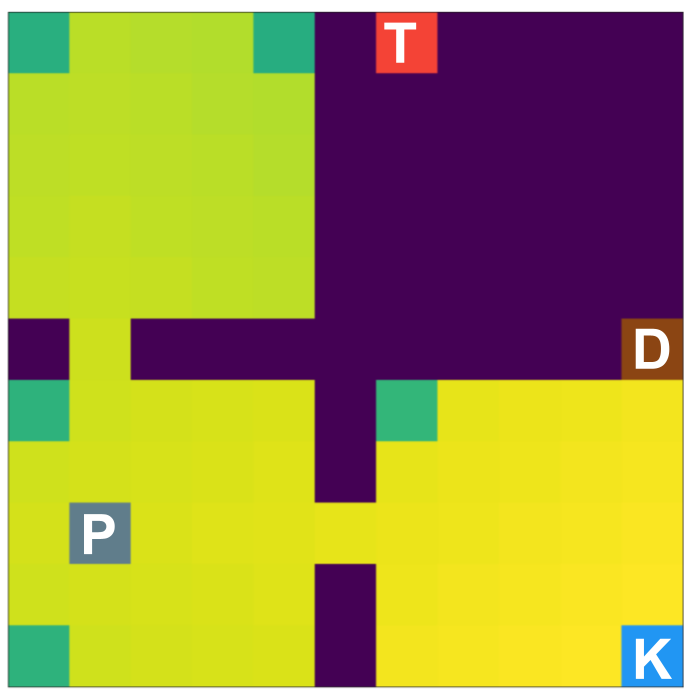} 
      \end{subfigure} 
      \begin{subfigure}[b]{0.24\linewidth}
        \centering
        \includegraphics[width=\linewidth]{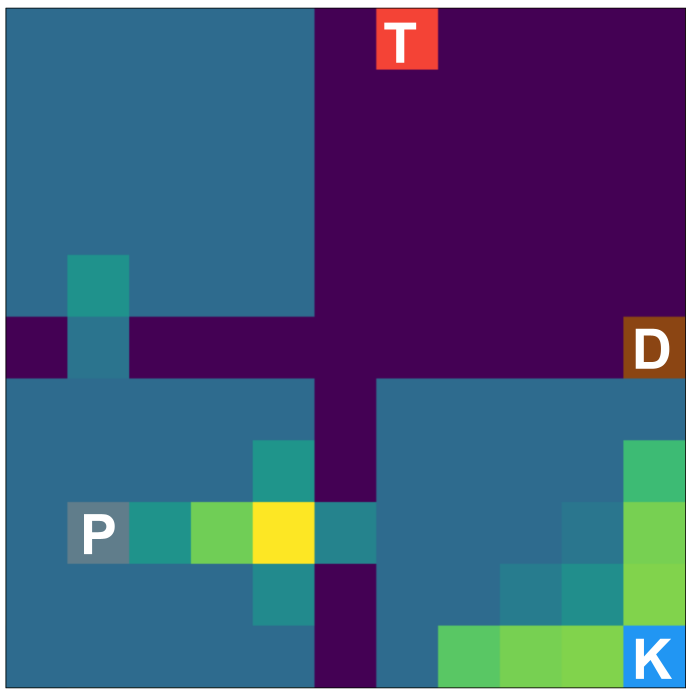}
      \end{subfigure}%
      \begin{subfigure}[b]{0.24\linewidth}
        \centering
        \includegraphics[width=\linewidth]{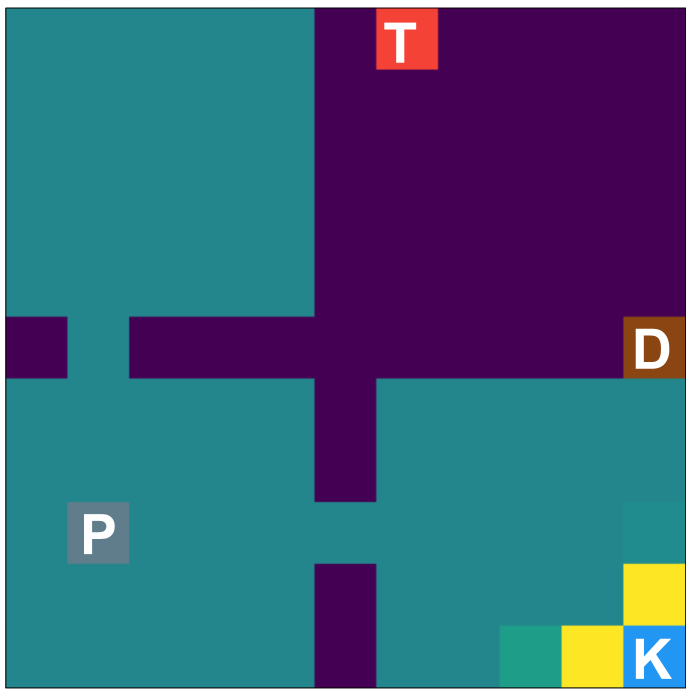}
      \end{subfigure} 
      \begin{subfigure}[b]{0.24\linewidth}
        \centering
        \includegraphics[width=\linewidth]{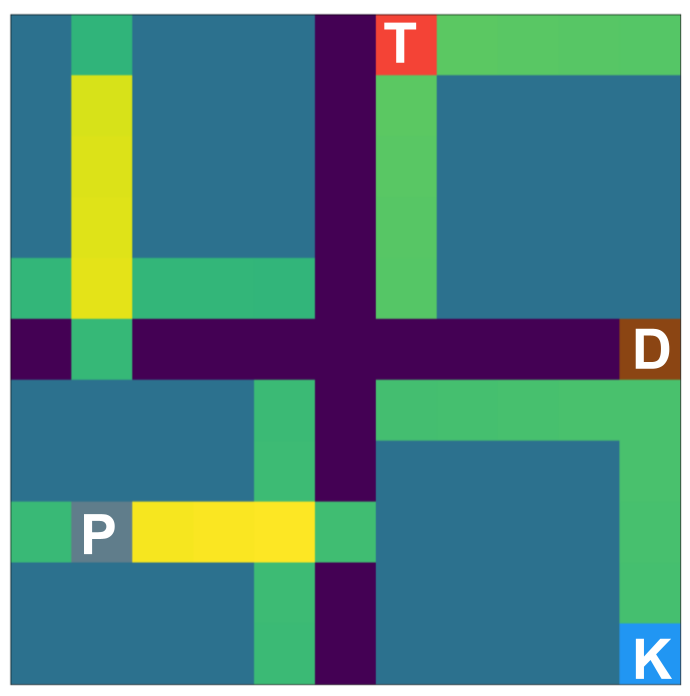}
        \caption{action gap}
      \end{subfigure}%
      \begin{subfigure}[b]{0.24\linewidth}
        \centering
        \includegraphics[width=\linewidth]{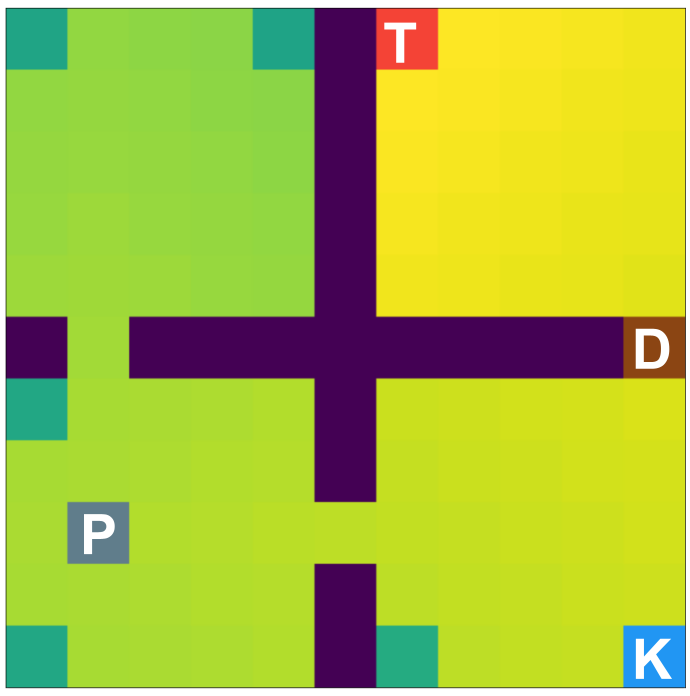} 
        \caption{important advice}
      \end{subfigure} 
      \begin{subfigure}[b]{0.24\linewidth}
        \centering
        \includegraphics[width=\linewidth]{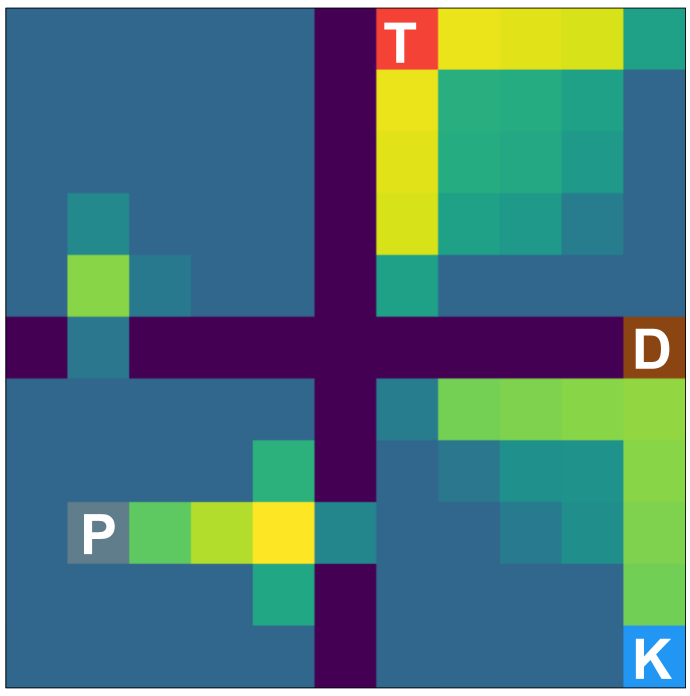}
        \caption{lazy-gap $\eta=0.03$}
      \end{subfigure}%
      \begin{subfigure}[b]{0.24\linewidth}
        \centering
        \includegraphics[width=\linewidth]{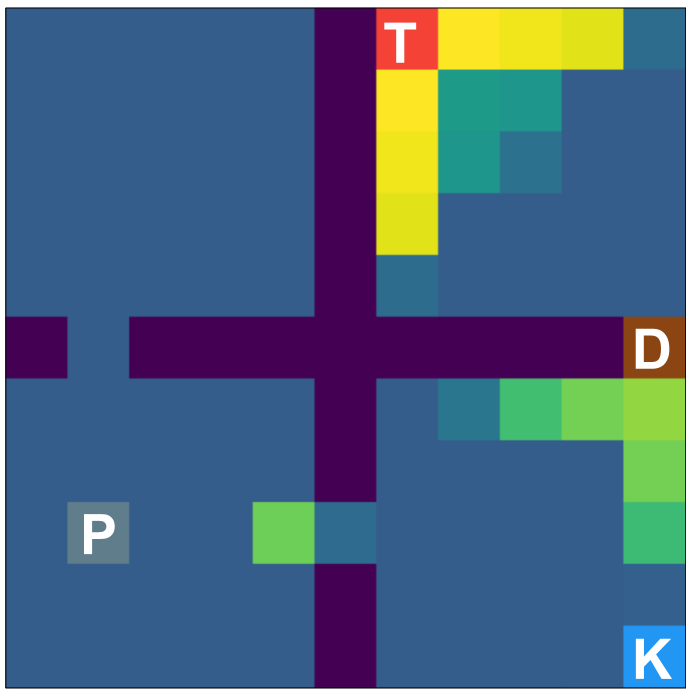}
        \caption{lazy-gap $\eta=0.05$}
      \end{subfigure} 
    \caption{State importance according to different metrics. Top row: KDT without the key. Bottom row: KDT with the key. (a) Importance according to the action-gap. (b) Importance according to importance advice. (c) Importance according to the lazy-gap with $\eta=0.03$. (d) Importance according to the lazy-gap with $\eta=0.05$. Brighter colors indicate higher importance.}
    \label{fig:importance}
\end{figure*}

Fig.~\ref{fig:importance} displays states importance according to these measures on the KDT environment. As visible, the lazy-gap only attributes importance to the states that lead to key actions (picking up the keys, passing through the doors, reaching the treasure). On the other hand, the action-gap uniformly emphasizes all the states along the trajectory of the optimal policy, while the importance advice is dominated by temporal proximity to the rewards and does not discriminate key actions.

\section{Atari curves}

\begin{figure*}[t]
    \centering
    \vspace{-1cm}
    \includegraphics[width=\textwidth]{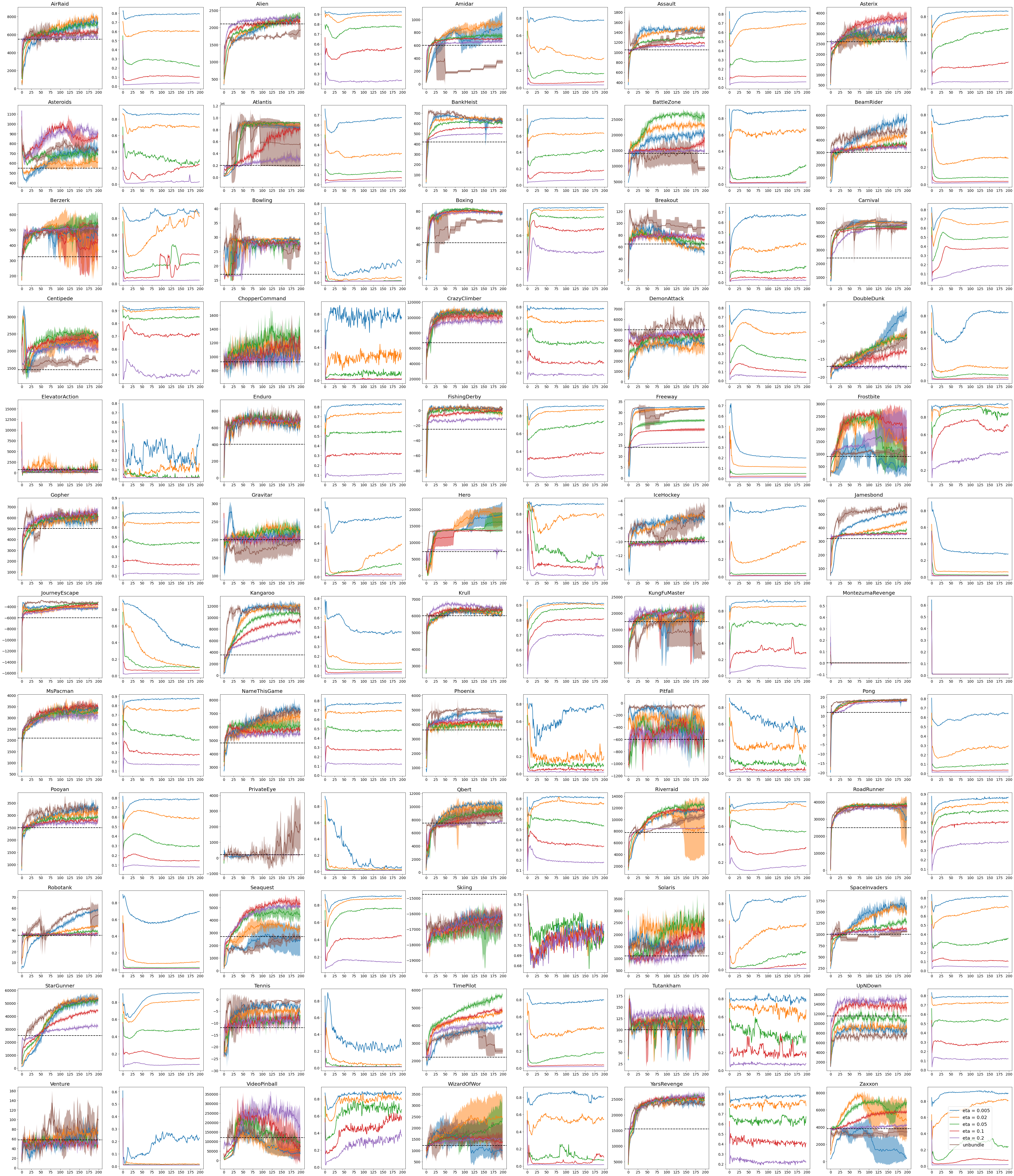}
    \caption{Performances on all 60 Atari 2600 games.}
    \label{fig:all-atari}
\end{figure*}

We display the in-game scores of DQN, DQN with warm restarts, and DQN in Lazy-MDPs in all Atari 2600 games that are mentioned in Sec.~\ref{subsec:residual} in Fig.~\ref{fig:all-atari}. For each game, two figures are provided: the left figure contains the scores, the right figure contains the fraction of controls (between 0 and 1, 0 being maximally lazy, and 1 taking control in all states) by the agent in the lazy-MDP.

\end{document}